\documentclass{article}
\usepackage[preprint]{log_2024}				% for anonymous submission to proceedings track

\usepackage{booktabs}						% professional-quality tables
\usepackage{multirow}						% tabular cells spanning multiple rows
\usepackage{amsfonts}						% blackboard math symbols
\usepackage{graphicx}						% figures
\usepackage{duckuments}						% sample images

\usepackage[numbers,compress,sort]{natbib}	% for numerical citations
\usepackage[utf8]{inputenc} % allow utf-8 input
\usepackage[T1]{fontenc}    % use 8-bit T1 fonts
\usepackage{url}            % simple URL typesetting
\usepackage{booktabs}       % professional-quality tables
\usepackage{amsfonts}       % blackboard math symbols
\usepackage{nicefrac}       % compact symbols for 1/2, etc.
\usepackage{microtype}      % microtypography
\usepackage{xcolor}         % colors

\usepackage{amsmath}
\usepackage{amssymb}
\usepackage{mathtools}
\usepackage{amsthm}
\theoremstyle{plain}
\newtheorem{theorem}{Theorem}[section]
\newtheorem{proposition}[theorem]{Proposition}

\newtheorem{corollary}[theorem]{Corollary}
\theoremstyle{definition}
\newtheorem{definition}[theorem]{Definition}

\theoremstyle{remark}

\newtheorem{proposition*}[theorem]{Proposition}
\newtheorem{theorem*}[theorem]{Theorem}

\usepackage{siunitx}

\usepackage{multirow}

\usepackage{subcaption}
\usepackage{tikz}
\usetikzlibrary{calc,arrows.meta,positioning,quotes}

\usepackage{pifont}

\title[Preventing Representational Rank Collapse in MPNNs by Splitting the Computational Graph]{Preventing Representational Rank Collapse in MPNNs by Splitting the Computational Graph}

\author[Roth et al.]{%
    Andreas Roth$^{1}$ \qquad Franka Bause$^{2,3}$ \qquad Nils M.~Kriege$^{2,4}$ \qquad Thomas Liebig$^{1,5}$ \\
        \small $^1$Faculty of Computer Science, TU Dortmund University, Dortmund, Germany \\  \email{\{andreas.roth,\,thomas.liebig\}@tu-dortmund.de} \\
 		\small $^2$Faculty of Computer Science, University of Vienna, Vienna, Austria \\ 
 		\small $^3$UniVie Doctoral School Computer Science, University of Vienna, Vienna, Austria \\
 		\small $^4$Research Network Data Science, University of Vienna, Vienna, Austria \\ \email{\{franka.bause,\,nils.kriege\}@univie.ac.at} \\
 		\small $^5$Lamarr Institute for Machine Learning and Artificial Intelligence, Dortmund, Germany
}

\begin{document}

\maketitle

\begin{abstract}
The ability of message-passing neural networks (MPNNs) to fit complex functions over graphs is limited as most graph convolutions amplify the same signal across all feature channels, a phenomenon known as rank collapse, and over-smoothing as a special case.
Most approaches to mitigate over-smoothing extend common message-passing schemes, e.g., the graph convolutional network, by utilizing residual connections, gating mechanisms, normalization, or regularization techniques. Our work contrarily proposes to directly tackle the cause of this issue by modifying the message-passing scheme and exchanging different types of messages using multi-relational graphs. We identify a sufficient condition to ensure linearly independent node representations. As one instantion, we show that operating on multiple directed acyclic graphs always satisfies our condition and propose to obtain these by defining a strict partial ordering of the nodes. We conduct comprehensive experiments that confirm the benefits of operating on multi-relational graphs to achieve more informative node representations.  
\end{abstract}

\section{Introduction}
\label{sec:intro}
Many challenging tasks, such as drug discovery~\citep{igashov2024equivariant}, social network predictions~\citep{fan2019graph}, and traffic prediction~\citep{derrow2021eta}, involve graph-structured data.
Message-passing neural networks (MPNNs)~\citep{gilmer2017neural} have found success in many of these areas. However, MPNNs did not see the same level of improvement against classical methods, such as graph kernels~\citep{NEURIPS2022_7eed2822}, that was achieved for computer vision~\citep{he2016deep} and natural language processing tasks~\citep{vaswani2017attention}.
MPNNs struggle to achieve satisfying performance for challenging tasks, such as large-scale heterophilic node classification.
Computational issues like over-smoothing~\cite{li2018deeper} and its more general form of representational rank collapse limit the ability of MPNNs to obtain informative node embeddings.
Rank collapse refers to the phenomenon that node representations become more similar in each feature dimension after each iteration of message-passing. %in the limit of infinitely many iterations results in linear dependece of all node representations. In the special case that these these vector correspond to low-frequency components, e.g., each feature is constant across nodes, this is referred to as over-smoothing.
Much attention has been paid to dealing with over-smoothing, e.g., by allowing either smooth signals or non-smooth signals to be amplified~\citep{yan2022two}. However, as MPNNs typically produce multi-dimensional features, models should be able to produce representations that are similar in one dimension but dissimilar in another dimension, e.g., when two people work together but pursue different hobbies. Few methods have yet to consider capturing both smooth and non-smooth features simultaneously. 
%one or two more sentences.

%Our work
In particular, it was identified that applying graph convolutions using a simple graph does not allow a different behavior across features~\citep{roth2023rank,roth2024simplifying}.
%we propose to split the edges of a graph into an multi-relational graph and operate MPNNs on these instead.
In this work, we propose splitting a given graph into a multi-relational graph and operating on multi-relational split MPNNs (MRS-MPNNs). Each edge is assigned a relation type, and messages are passed using distinct feature transformations before the results are combined to form a single state within each graph convolution.
Our theoretical analysis provides sufficient conditions to ensure more informative node representations. Specifically, different structural properties between edge relations are required.
While this theory opens many different directions for constructing multiple graphs that satisfy our conditions, we propose an instantiation that forms directed acyclic graphs (DAGs) with each relation type. These are constructed by defining a strict partial ordering over the nodes and assigning each edge to a relation type depending on the ordering between the two nodes.
%This leads to MRS-MPNNs, for which messages are processed on each computational graph using distinct feature transformations and then aggregated into a joint state.
%This framework can be directly combined with any MPNN.
%Experiments
In our experiments, we evaluate several choices for the partial order and identify the node degree as a powerful and general choice.
We empirically confirm that MRS-MPNN prevents rank collapse and that its ability to amplify different signals for each feature column benefits the learning process for several methods.
%When the DA framework is combined with Dir-GNN, a state-of-the-art method for node classification on heterophilic directed graphs, we further improve their results on all five considered datasets. 
We summarize our main contributions as follows:
\begin{itemize}
\item We propose to split the edges of graphs in multiple edge relations and utilize multi-relational split MPNNs (MRS-MPNNs). We establish the necessary and sufficient condition on the edges of each relation type so that node representations become more informative (Section~\ref{sec:theory}).
\item As one instantiation that satisfies our condition, we propose to utilize multiple edge relations that are directed and acyclic for which define a strict partial ordering of the nodes. Each edge is assigned a relation type according to the ordering of its adjacent nodes (Section~\ref{sec:method}).
%    \item We show that the ergodicity of a graph leads to over-smoothing in MPNNs, and that graphs without ergodic components do not over-smooth. Our theory further shows the benefits of operating on multiple DAGs as this also avoids representational rank collapse (Section~\ref{sec:theory}).
%    \item We propose MRS-MPNNs, a framework that splits any graph into three computational graphs based on a strict partial ordering of the nodes. This framework can be applied to any MPNN by constructing different messages depending on the order between nodes (Section~\ref{sec:method}).
    \item Our experiments confirm our theory by demonstrating that MRS-MPNNs prevent rank collapse and improve the learning process (Section~\ref{sec:exp}).
\end{itemize}

\section{Preliminaries}
\label{sec:preliminaries}
Let $G = (\mathcal{V}, \mathcal{E})$ be a (simple) graph, where $\mathcal{V} = \{v_1,\dots, v_n\}$ is its set of $n$ nodes and $\mathcal{E}\subseteq \mathcal{V}\times \mathcal{V}$ its set of edges. We refer to $\mathbf{A}\in\mathbb{R}^{n\times n}$ as the adjacency matrix for which $\mathbf{A}_{ij} = 1$ if $(v_i,v_j)\in \mathcal{E}$, otherwise the entry is $0$. The nodes with an edge ending at $v_i$ are defined as its neighbors $N_i = \{ v_k \mid (v_k,v_i)\in \mathcal{E}\}$. 
Based on each node's incoming edges, we define the degree matrix $\mathbf{D}\in\mathbb{R}^{n\times n}$ as a diagonal matrix with $d_{ii} = |N_i|$. A multi-relational graph $(\mathcal{V},\mathcal{E}_1,\dots,\mathcal{E}_l)$ is a graph that contains multiple relations or edge types.

\paragraph{Message-Passing Neural Networks}
Given a graph $G$ and $d$-dimensional features $\mathbf{X}\in\mathbb{R}^{n\times d}$ for each node, message-passing neural networks (MPNNs) aim to obtain informative node representations capturing both structural properties and connections between node features. Most MPNNs follow an iterative message-passing scheme that updates each node's representations 
\begin{equation}
\label{eq:mpnn}
    \mathbf{x}^{\prime}_i = \phi\left(\mathbf{x}_i,\bigoplus_{j\in N_i}\psi\left(\mathbf{x}_i,\mathbf{x}_j\right)\right)
\end{equation}
using a message function $\psi$, a permutation invariant aggregation function $\oplus$, and a combine function $\phi$. This is typically repeated for $k$ instantiations. 
State-of-the-art MPNNs add additional components such as residual connections~\citep{chen2020simple,scholkemper2024residual}, restart terms~\citep{chen2020simple,roth2022transforming}, or gating mechanisms~\citep{li2015gated,rusch2022gradient}.
However, for exchanging messages, most methods follow a simple scheme that can be expressed in matrix notation
\begin{equation}
\label{eq:message}
    \bigoplus_{j\in N_i}\psi\left(\mathbf{x}_i,\mathbf{x}_j\right) = \left[\mathbf{\Tilde{A}}\mathbf{X}\mathbf{W}\right]_i
\end{equation}
where $\mathbf{\Tilde{A}}\in\mathbb{R}^{n\times n}$ corresponds to the aggregation function. $\mathbf{\Tilde{A}}$ may be the symmetrically normalized adjacency matrix, the mean aggregation, the sum aggregation, or contain negative values. It may also include self-loops. Most models apply a linear feature transformation $\mathbf{W}\in\mathbb{R}^{d\times d^\prime}$.

\paragraph{Rank Collapse and Over-Smoothing}
For many MPNNs, node representations tend to become more similar as more update iterations are performed, limiting the learnable functions over graphs.
All graph convolutions of the form given by Eq.~\ref{eq:message} were found to amplify and damp the same signals across all features~\citep{roth2023rank,roth2024simplifying}. While the features for two adjacent nodes can get closer to each other or further apart, this behavior is the same for all features. 
In the limit, all node representations become linearly dependent, resulting in a rank-one matrix. 
This phenomenon limits the information that can be present in representations and is referred to as rank collapse~\citep{roth2023rank}.
%MPNNs of this form limit the information that can be obtained. 
%The dominating signal is given by the dominating eigenvector of $\mathbf{\Tilde{A}}$~\citep{li2018deeper,cai2020note,digiovanni2023understanding,roth2023rank}. This phenomenon is referred to as over-correlation~\citep{TODO} or representational rank collapse~\citep{roth2023rank}. 
As a special case of rank collapse, over-smoothing occurs when representations become linearly dependent and contain smooth values~\citep{li2018deeper,oono2019graph,cai2020note,digiovanni2023understanding}.
%To determine how close representations are to a rank-one matrix, the rank-one distance (ROD) is defined as
%\begin{equation}
%\mathrm{ROD}(\mathbf{X}) = \left\| \frac{\mathbf{X}}{\|\mathbf{X}\|} - \frac{\mathbf{uv}^T}{\|\mathbf{uv}^T\|}\right\|
%\end{equation}
%where $\mathbf{u}\in\mathbb{R}^n$ is the column $\mathbf{v}\in\mathbb{R}^d$ the row of $\mathbf{X}$ with the largest norm~\citep{roth2024simplifying}. As this metric generalizes the Dirichlet energy~\citep{cai2020note}, constructing models that keep ROD constant, also prevent over-smoothing.

\section{Related Work}

\paragraph{Dealing with Rank Collapse and Over-Smoothing}
%Many methods aim to prevent over-smoothing
Various methods aim to reduce over-smoothing in MPNNs. 
These include combining the output of Eq.~\ref{eq:message} with previous states, e.g., by utilizing residual connections~\citep{bresson2017residual,chen2020simple,scholkemper2024residual} or restart terms~\citep{gasteiger2018combining,chen2020simple,roth2022transforming}.
Gating mechanisms were proposed to stop updating node states after varying numbers of iterations~\citep{rusch2022gradient,finkelshtein2023cooperative}.
Normalization operations that reduce the similarity between representations were introduced~\citep{Zhao2020PairNorm,li2020deepergcn}. Another line of research proposes regularization terms to punish smooth representations during optimization~\citep{zhou2021dirichlet}.

Several works study the construction of MPNNs that can amplify different signals, e.g., either low-frequency or high-frequency signals~\citep{eliasof2023improving,digiovanni2023understanding}, or mixtures using negative edge weights~\citep{bo2021beyond, yan2022two} or combining multiple aggregation functions~\citep{corso2020principal,tailor2022adaptive}.
It is still unclear how a single graph convolution can amplify different signals for each feature channel, as few methods have been proposed to mitigate the more general rank collapse. \citet{jin2022feature} propose to regularize the correlation between node representations.
All of these methods utilize the base message-passing scheme (Eq.~\ref{eq:message}) using a single edge relation, which is known to suffer from rank collapse for almost every $\Tilde{\mathbf{A}}$ and $\mathbf{W}$~\citep{roth2023rank,roth2024simplifying}.

\paragraph{Changing the Computational Graphs}
%Disconnecting the Computational Graph from the Input Graph
Typically, MPNNs operate directly on the input graph $G = (\mathcal{V},\mathcal{E})$, which may cause computational issues like over-squashing~\citep{alon2021on} as information cannot flow over large distances.
Several methods propose to perform the message-passing on a computational graph $G^\prime = (\mathcal{V},\mathcal{E}^\prime)$ that has better-suited structural properties, e.g., by graph rewiring techniques~\citep{topping2022understanding,abboud2022shortest,barbero2024localityaware}. Co-GNNs modify the computational graph in each step by allowing each node to choose between sending messages, listening, or isolating~\citep{finkelshtein2023cooperative}.
Dense connectivity, including higher node degrees~\citep{yan2022two} and a larger curvature~\citep{pmlr-v202-nguyen23c} were identified to amplify over-smoothing. However, structural properties leading to rank collapse and how to construct beneficial computational graphs remain unclear. 

%Specific graph structures
%Other methods were proposed for specific graph structures, e.g., for trees~\citep{pmlr-v37-zhub15,kiperwasser2016easy} or directed acyclic graphs~\citep{shuai2016dag,zhang2019d,thost2021directed}. However, these methods do not consider computational benefits of specific graph structures or apply to graphs that are not given in that form.

Various MPNNs for multi-relational graphs have been proposed~\cite{schlichtkrull2018modeling,Vashishth2020Composition}. However, these assume multiple relations to be given in the data. %It is unclear whether learning on multi-relational graphs is desired for simple graphs and how it relates to phenomena like rank collapse and over-smoothing.
Other methods explored forming multiple computational graphs. \citet{suresh2021breaking} add computational graphs that connect nodes with a large structural similarity to account for both proximity and structural similarity with different edge types. Factorizable graph convolutional networks disentangle edges of a graph into multiple interpretable factor graphs to produce disentangled representations~\citep{yang2020factorizable}. ES-GNN~\citep{guo2022esgnn} learns to split the edges of a graph into two sets, one containing task-relevant edges, and one containing task-irrelevant edges. Predictions are obtained by performing message-passing using the task-relevant edges. EXPASS~\citep{giunchiglia2022towards} modifies the computational graph by weighting edges so that edge weights of low importance based on an explanation method are reduced. ACM~\citep{luan2022revisiting} uses both the graph Laplacian and normalized adjacency matrices as computational graphs to amplify low-frequency and high-frequency signals.
ADR-GNNs~\citep{eliasof2023adrgnn} learn an advection-diffusion-reaction system for which edge weights are learned channel-wise independently, but a shared transformation is applied.
For directed graphs, Dir-GNNs add the reverse direction as a second computational graph~\citep{rossi2023edge}. 
Subgraph GNNs~\citep{bevilacqua2022equivariant,bevilacqua2024efficient} similarly generate multiple subgraphs based on a given policy. Message-passing is performed separately for each subgraph, which can thus lead to representational rank collapse on each subgraph. 
Despite these works, the general benefits of operating on multiple computational graphs and how these lead to more informative node representations are still unclear. Studying the potential benefits and required properties of multiple computational graphs can help us better understand the advantages of all these methods.

%\paragraph{Operating on Directed Acyclic Graphs}
%The given graph may already satisfy specific properties. Related to our work, several methods for operating on trees and on DAGs were proposed.
%Datasets satisfying these properties are rare. These methods' main application areas are source code and neural architectures~\citep{thost2021directed}. As the given graph is typically not a DAG, these methods are not broadly applicable. The connection between the computational benefits of these data structures has not been studied.
\section{Splitting MPNNs into Multi-Relational MPNNs}
As each iteration of message-passing on a simple graph makes feature columns more similar~\cite{roth2023rank,roth2024simplifying}, we study the effects of utilizing multiple edge relations by splitting a graph into a multi-relational graph. The goal is to be able to obtain more informative embeddings by allowing for features to become more similar in one dimension while becoming more dissimilar in another dimension. 
Formally, we let $l$ be the number of relation types we want to split our graph into. For each edge $(v_i,v_j)\in\mathcal{E}$, we apply a permutation invariant edge relation assignment function $f\colon (v_i,v_j) \mapsto \{1,\dots,l\}$. Note that while we consider assigning each edge to a single relation type, multiple assignments and continuous assignment scores are also covered by our following theory. Based on this assignment function, one of the relation types and corresponding feature transformations $\psi_1,\dots,\psi_l$ is selected for each edge. In general, we define one iteration of MRS-MPNNs as follows:

\begin{definition}(Multi-Relational Split MPNNs (MRS-MPNNs))
\begin{equation}
\begin{split}
    \mathbf{x}^{\prime}_i 
    &=  \phi\left(\mathbf{x}_i,\bigoplus_{j\in N_i}\psi_{f(v_i,v_j)}\left(\mathbf{x}_i,\mathbf{x}_j\right)\right)\, ,
    \end{split}
\end{equation}
where $\bigoplus$ and $\phi$ are an aggregation function and a combination function as used in an MPNN. 
\end{definition}
Thus, all MPNNs can be transformed into an MRS-MPNN by duplicating the message function and defining an edge relation assignment function $f$. As an example, we state the graph convolutional network (GCN)~\citep{kipf2016semi} in the MRS framework, as it is commonly used and often serves as the message-passing component within complex models.

\paragraph{MRS-GCN}
Given some node representations $\mathbf{X}^{n\times d}$, we define the MRS-GCN as
\begin{equation}
\begin{split}
    \left[\textrm{MRS-GCN}\left(\mathbf{X}, \mathcal{E}, f\right)\right]_i &\coloneq \left[\Tilde{\mathbf{A}}_1\mathbf{X}\mathbf{W}_1 + \dots + \Tilde{\mathbf{A}}_l\mathbf{X}\mathbf{W}_l\right]_i \\ 
    &=  \sum_{j\in N_i} \frac{1}{\sqrt{d_i}\sqrt{d_j}} \mathbf{W}_{f(v_i,v_j)}\mathbf{x}_j,
\end{split}
\end{equation}
where $d_i$ is the degree of node $i$, $\mathbf{W}_{1},\dots,\mathbf{W}_{l}\in\mathbb{R}^{d\times d^\prime}$ are feature transformations and $\Tilde{\mathbf{A}}_1,\dots,\Tilde{\mathbf{A}}_l\in\mathbb{R}^{n\times n}$ contain the edge weights of the corresponding computational graph, i.e., $\Tilde{\mathbf{A}}_1+\dots+\Tilde{\mathbf{A}}_3 = \mathbf{D}^{-1/2}\mathbf{A}\mathbf{D}^{-1/2}$. Compared to the GCN, only the selected transformation $\mathbf{W}_{f(v_i,v_j)}$ changes. We define additional models in their MRS form in Appendix~\ref{sec:model_details}.

\paragraph{Theoretical Properties}
\begin{figure*}
\definecolor{viridis1}{RGB}{72,21,103}
\definecolor{viridis2}{RGB}{40,125,142}
\definecolor{viridis3}{RGB}{115,208,85}
     \begin{subfigure}[b]{0.3\textwidth}
         \centering
         
         \begin{tikzpicture}
\begin{scope}[every node/.style={circle,draw,scale=1.6}]
    \node (D) at (1.5,1.5) {1} ;
    \node (E) at (3.5,1.5) {2} ;
\end{scope}

\begin{scope}[>={Stealth[viridis1]},
              every edge/.style={draw=viridis1,thick}]
    \path [->] (0.5,3.0) edge["1"] (D); 
    \path [->] (1.16,3.0) edge["1"] (D); 
    \path [->] (1.88,3.0) edge["1",pos=0.1] (D); 
    \path [->] (2.5,3.0) edge["1"] (D); 
    \path [->] (3.0,3.0) edge["1"] (E); 
    \path [->] (4.0,3.0) edge["1"] (E); 
\end{scope}
\begin{scope}[>={Stealth[viridis3]},
              every edge/.style={draw=viridis3,thick}]
    \path [->] (1.0,0.0) edge["3"] (D); 
    \path [->] (2.0,0.0) edge["-1"] (D);
    \path [->] (3.5,0.0) edge["1"] (E); 
\end{scope}
\end{tikzpicture}
        
         \caption{Structurally dependent nodes.}
     \end{subfigure}
     \hfill
\begin{subfigure}[b]{0.33\textwidth}
         \centering
         
         \begin{tikzpicture}
\begin{scope}[every node/.style={circle,draw,scale=1.6}]
    \node (D) at (1.5,1.5) {3} ;
    \node (E) at (3.5,1.5) {4} ;
\end{scope}

\begin{scope}[>={Stealth[viridis1]},
              every edge /.style={draw=viridis1,thick},
              ]
    \path [->] (0.5,3.0) edge["1"] (D); 
    \path [->] (1.5,3.0) edge["1"] (D); 
    \path [->] (2.5,3.0) edge["1"] (D); 
    \path [->] (3.0,3.0) edge["1"] (E); 
    \path [->] (4.0,3.0) edge["1"] (E); 
\end{scope}
\begin{scope}[>={Stealth[viridis3]},
              every edge/.style={draw=viridis3,thick}]
    \path [->] (1.0,0.0) edge["1"] (D); 
    \path [->] (2.0,0.0) edge["1"] (D);
    \path [->] (3.5,0.0) edge["1"] (E); 
\end{scope}
\end{tikzpicture}
        
         \caption{Structurally independent nodes.}
     \end{subfigure}
     \hfill
\begin{subfigure}[b]{0.32\textwidth}
         \centering
         
         \begin{tabular}{cccc} \toprule
  Node & $d_1$ & $d_2$ & SD \\ \midrule
  1 & 4 & 2 & \ding{51} \\
  2 & 2 & 1 & \ding{51} \\
  3 & 3 & 2 & \ding{55} \\
  4 & 2 & 1 & \ding{51} \\
  \bottomrule
  \end{tabular}
        
         \caption{Numerical dependency.}
     \end{subfigure}
        \caption{Structurally dependent (a) and independent (b) node pairs. Colors indicate different graphs. Numbers beside edges indicate edge weights. SD refers to the structural dependency of nodes based on $d_1$ and $d_2$ (given by Def.~\ref{def:neigh_struc}).}
        \label{fig:struct_independ}
\end{figure*}

We now show the computational benefits of operating on multiple relations. We do that by identifying a sufficient condition on the structure of the edge sets, that ensures linear independence of node representations.
In this analysis, we consider instantiations of MRS-MPNN of the form 
\begin{equation}
    \mathbf{X}^\prime = \sigma(\mathbf{A}_1\mathbf{X}\mathbf{W}_1 + \dots + \mathbf{A}_l\mathbf{X}\mathbf{W}_l)
\end{equation}
where $\mathbf{X}\in\mathbb{R}^{n\times d}$ are node representations, $\mathbf{A}_1,\dots,\mathbf{A}_l\in\mathbb{R}^{n\times n}$ represent computational graphs, and $\mathbf{W}_1,\dots,\mathbf{W}_l\in\mathbb{R}^{d\times d^\prime}$ are feature transformations. Note, that MRS-GCN and MRS-SAGE are special cases of this form. Our analysis requires the weighted in-degree of each node and graph: 
\label{sec:theory}
\begin{definition}(Weighted in-degrees)
\label{def:neigh_struc}
 Let $\mathbf{A}_1,\dots,\mathbf{A}_l\in\mathbb{R}^{n\times n}$ represent $l$ edge relations with any edge weights. For each node $i\in[n]$, the vector of weighted in-degrees is defined as 
 \begin{equation}
 \mathbf{d}^i = \begin{bmatrix} d^i_1 & \dots & d^i_l\end{bmatrix}\, ,
 \end{equation}
 where $d^i_k = \sum_{m\in[n]} \mathbf{A}_k[i,m]$ is the weighted in-degree of node $i$ in edge relation $k$.
\end{definition}

With this, we introduce the concept of structural dependence of a pair of nodes as the linear dependence of their weighted in-degrees:

\begin{definition} (Structural dependence and independence)
    Let $\mathbf{A}_1,\dots,\mathbf{A}_l\in\mathbb{R}^{n\times n}$ be matrices and $\mathbf{d}^i = \begin{bmatrix} d^i_1 & \dots & d^i_l\end{bmatrix}\in\mathbb{R}^l$ be the vector of weighted in-degrees for $i\in[n]$.
    A pair of nodes $v_i,v_j$ is said to be \textbf{structurally independent} if the vectors $\mathbf{d}^i$ and $\mathbf{d}^j$ are linearly independent. Otherwise, they are called \textbf{structurally dependent}.
\end{definition}

We provide an example in Figure~\ref{fig:struct_independ}.
This serves as a sufficient condition that ensures that node pairs get mapped to linearly independent representations: 

\begin{theorem}\label{theorem:node} (Structurally independent nodes produce linearly independent representations.)
    Let $\mathbf{A}_1,\dots,\mathbf{A}_l\in\mathbb{R}^{n\times n}$ be $l$ matrices with nodes $v_i,v_j$ being structurally independent. Then,
        \begin{align}
    \mathbf{x}_i^\prime = \left[\mathbf{A}_1\mathbf{X}\mathbf{W}_1 + \dots + \mathbf{A}_l\mathbf{X}\mathbf{W}_l\right]_i, \\
    \mathbf{x}_j^\prime = \left[\mathbf{A}_1\mathbf{X}\mathbf{W}_1 + \dots + \mathbf{A}_l\mathbf{X}\mathbf{W}_l\right]_j,
    \end{align}
    are linearly independent for a.e. $\mathbf{W}_1,\dots,\mathbf{W}_l\in\mathbb{R}^{d\times d^\prime}$ with $d,d^\prime > l \geq 1$ and a.e. $\mathbf{X}\in\mathbb{R}^{n\times d}$ with $\mathrm{rank}(\mathbf{X}) = 1$
\end{theorem}

We provide all proofs in Appendix~\ref{sec:proofs}. Linearly dependent node representations do not form fixed points or invariant subspaces for two structurally independent nodes. 
This makes it impossible for two such nodes to converge to overly similar representations. 
Node representations can get more similar in one feature dimension and more dissimilar in another dimension, allowing representations to capture more information.
Also note that for component-wise injective activation functions $\sigma$, we also know that $\sigma(\mathbf{x}_i)$ is linearly independent to $\sigma(\mathbf{x}_j)$ for a.e. vectors $\mathbf{x}_i$, $\mathbf{x}_j$. Another intuitive explanation comes from connecting structurally independent nodes to their unfolding trees~\citep{sato2021random}: These nodes always have different unfolding trees, and MRS-MPNNs will assign not just different but linearly independent representations.

Consequently, rank collapse, and therefore over-smoothing, are prevented when operating on multiple graphs with structurally independent nodes. As a side note, we also highlight the connection to the expressivity of an MPNN. Both phenomena have the goal of preventing node states from becoming equal when their inputs are different. Two structurally independent nodes always get distinguished. Thus, the more structurally independent nodes exist in a graph, the more nodes are distinguished. 

%\begin{definition} (Set of structurally independent nodes.)
%    A set of nodes $\mathcal{V}_l = \{v_1,\dots,v_l\}$ is said to be structurally independent on a set of graphs $\mathcal{A}_l = \{\mathbf{A}_1,\dots,\mathbf{A}_l\}$ if for all nodes $v_q\in\mathcal{V}_l$, there do not exist scalars $c_2,\dots,c_l\in\mathbb{R}$ such that for all graphs $\mathbf{A}_p\in\mathcal{A}_l$, the edge weight sum of node $v_q$ is a linear combination of the edge weight sums of the other nodes, i.e., 
%    \begin{equation}
%        d_p[q] = \sum_{v_r\in\mathcal{V}_l\setminus v_q} c_r d_p[r].
%    \end{equation}
%\end{definition}

%\begin{theorem} (Functional Expressivity)
%    Let $\mathbf{X}\in\mathbb{R}^{n\times d}$, $\mathbf{A}_1,\dots,\mathbf{A}_k\in\mathbb{R}^{n\times n}$ be $k$ adjacency matrices, and $\mathbf{W}^{(l)},\dots\mathbf{W}^{(k)}\in\mathbf{R}^{d\times d^\prime}$ be feature transformations with $n\geq d^\prime\geq k$.
%    Then, $rank(\sum_{l=1}^k\mathbf{A}^{(l)}\mathbf{X}\mathbf{W}^{(l)}) \geq l$, where $l\leq k$ is the size of the largest set of structurally independent nodes.
%\end{theorem}

We are interested not only in this structural expressivity but also in the information stored in the node representations. 

\begin{theorem} (Structural independence prevents rank collapse.)
    Given $n$ nodes, let $\mathbf{E} = \begin{bmatrix} \mathbf{d}^1 & \dots & \mathbf{d}^n\end{bmatrix}\in\mathbb{R}^{n\times l}$ be the matrix of weighted in-degrees. Let $\sigma$ be a component-wise injective activation function.
    Then, for a.e. $\mathbf{W}_1,\dots\mathbf{W}_l\in\mathbf{R}^{d\times d^\prime}$ with $n\geq d^\prime\geq l$, 
    \begin{equation}
    \mathrm{rank}(\sigma(\sum_{l=1}^k\mathbf{A}_l\mathbf{X}\mathbf{W}_l)) \geq l,
    \end{equation}
    where $l = \mathrm{rank}(\mathbf{E})$ for a.e. $\mathbf{X}\in\mathbb{R}^{n\times d}$. 
\end{theorem}

We have now confirmed that the minimal rank of representations gets larger with more structurally independent nodes. 

%\subsection{Computational Complexity}
%One concern of introducing multiple edge relations is the computational cost. Collecting and aggregating messages are the main computational costs of MPNNs, which remain unchanged when only applying different feature transformation for each message. MRS-MPNNs construct the same number of messages, with each message having the same feature dimension and utilizing the same number of parameters. The main computational difference is the graph splitting or construction for which we discuss one option in the following section.
\section{Obtaining Multiple Relations using a Partial Ordering}
\label{sec:method}

\begin{figure*}
\definecolor{viridis1}{RGB}{72,21,103}
\definecolor{viridis2}{RGB}{40,125,142}
\definecolor{viridis3}{RGB}{115,208,85}
     \centering
     \begin{subfigure}[b]{0.21\textwidth}
         \centering
        \resizebox{\textwidth}{\textwidth}{
         \begin{tikzpicture}
\begin{scope}[every node/.style={circle,draw,scale=1.0}]
    \node (A) at (0.5,3) {};
    \node (B) at (2.5,3) {};
    \node (C) at (0,1.5) {};
    \node (D) at (1,1.5) {} ;
    \node (E) at (2,1.5) {} ;
    \node (F) at (3,1.5) {} ;
    \node (G) at (0.5,0) {} ;
    \node (H) at (1.5,0) {} ;
\end{scope}

\begin{scope}[>={Stealth[black]},
              every edge/.style={draw=black,thick}]
    \path [<->] (A) edge (B);
    \path [<->] (A) edge (C);
    \path [<->] (A) edge (E);
    \path [<->] (A) edge (G);
    \path [<->] (B) edge (D);
    \path [<->] (B) edge (E); 
    \path [<->] (B) edge (F); 
    \path [<->] (C) edge (G); 
    \path [<->] (D) edge (H); 
    \path [<->] (E) edge (H); 
    \path [<->] (C) edge (D); 
\end{scope}
\end{tikzpicture}
        }
         \caption{Original graph.}
     \end{subfigure}
     \hfill
     \begin{subfigure}[b]{0.21\textwidth}
         \centering
         \resizebox{\textwidth}{\textwidth}{
         \begin{tikzpicture}
\begin{scope}[every node/.style={circle,draw,scale=1.0}]
    \node (A) at (0.5,3) {};
    \node (B) at (2.5,3) {};
    \node (C) at (0,1.5) {};
    \node (D) at (1,1.5) {} ;
    \node (E) at (2,1.5) {} ;
    \node (F) at (3,1.5) {} ;
    \node (G) at (0.5,0) {} ;
    \node (H) at (1.5,0) {} ;
\end{scope}

\begin{scope}[>={Stealth[viridis1]},
              every edge/.style={draw=viridis1,thick}]
    \path [->] (A) edge (C);
    \path [->] (A) edge (E);
    \path [->] (A) edge (G);
    \path [->] (B) edge (D);
    \path [->] (B) edge (E); 
    \path [->] (B) edge (F); 
    \path [->] (C) edge (G); 
    \path [->] (D) edge (H); 
    \path [->] (E) edge (H); 
\end{scope}
\end{tikzpicture}
        }
         \caption{DAG.}
     \end{subfigure}
     \hfill
     \begin{subfigure}[b]{0.21\textwidth}
         \centering
         \resizebox{\textwidth}{\textwidth}{
         \begin{tikzpicture}
\begin{scope}[every node/.style={circle,draw}]
    \node (A) at (0.5,3) {};
    \node (B) at (2.5,3) {};
    \node (C) at (0,1.5) {};
    \node (D) at (1,1.5) {} ;
    \node (E) at (2,1.5) {} ;
    \node (F) at (3,1.5) {} ;
    \node (G) at (0.5,0) {} ;
    \node (H) at (1.5,0) {} ;
\end{scope}

\begin{scope}[>={Stealth[viridis2]},
              every edge/.style={draw=viridis2,thick}]
    \path [<-] (A) edge (C);
    \path [<-] (A) edge (E);
    \path [<-] (A) edge (G);
    \path [<-] (B) edge (D);
    \path [<-] (B) edge (E); 
    \path [<-] (B) edge (F); 
    \path [<-] (C) edge (G); 
    \path [<-] (D) edge (H); 
    \path [<-] (E) edge (H); 
\end{scope}
\end{tikzpicture}
        }
         \caption{Reverse DAG.}
     \end{subfigure}
     \hfill
     \begin{subfigure}[b]{0.21\textwidth}
         \centering
         \resizebox{\textwidth}{\textwidth}{
         \begin{tikzpicture}
\begin{scope}[every node/.style={circle,draw,scale=1.0}]
    \node (A) at (0.5,3) {};
    \node (B) at (2.5,3) {};
    \node (C) at (0,1.5) {};
    \node (D) at (1,1.5) {} ;
    \node (E) at (2,1.5) {} ;
    \node (F) at (3,1.5) {} ;
    \node (G) at (0.5,0) {} ;
    \node (H) at (1.5,0) {} ;
\end{scope}

\begin{scope}[>={Stealth[viridis1]},
              every edge/.style={draw=viridis1,thick,bend right=10}]
    \path [->] (A) edge (C);
    \path [->] (A) edge (E);
    \path [->] (A) edge (G);
    \path [->] (B) edge (D);
    \path [->] (B) edge (E); 
    \path [->] (B) edge (F); 
    \path [->] (C) edge (G); 
    \path [->] (D) edge (H); 
    \path [->] (E) edge (H); 
\end{scope}
\begin{scope}[>={Stealth[viridis2]},
              every edge/.style={draw=viridis2,thick,bend left=10}]
    \path [<-] (A) edge (C);
    \path [<-] (A) edge (E);
    \path [<-] (A) edge (G);
    \path [<-] (B) edge (D);
    \path [<-] (B) edge (E); 
    \path [<-] (B) edge (F); 
    \path [<-] (C) edge (G); 
    \path [<-] (D) edge (H); 
    \path [<-] (E) edge (H); 
\end{scope}
\begin{scope}[>={Stealth[viridis3]},
              every edge/.style={draw=viridis3,thick}]
    \path [<->] (A) edge (B); 
    \path [<->] (C) edge (D); 
\end{scope}
\end{tikzpicture}
        }
         \caption{$\textrm{MRS-MPNN}_{\textrm{DA}}$.}
     \end{subfigure}
        \caption{Different computational graphs for $\textrm{MRS-MPNN}_{\textrm{DA}}$.}
        \label{fig:comp_graphs}
\end{figure*}

A suitable edge relation assignment function $f\colon (v_i,v_j) \mapsto \{1,\dots,l\}$ can be constructed in various ways. These include application-specific graph assignments, e.g., different relations for specific atom pairs in molecular graphs, data-dependent assignments, e.g., depending on initial or intermediate features, or assignments based on structural graph properties.

We derive one method based on observing that attentional message-passing, e.g., graph attention network~\citep{velivckovic2018graph} or graph transformers~\citep{shi2021masked}, utilize multiple attention heads, which can be seen as multiple edge relation type. For each attention head or relation type, the attention scores are normalized using the softmax function, i.e., each node aggregates its neighbors using a weighted mean for aggregation on each relation. 
These are typically applied to ergodic graphs or relations, which refer to them being strongly connected and aperiodic, i.e.,  a path exists between each pair of nodes, and the lengths of its cycles do not have a common divisor $> 1$. This is a typical assumption that is also used for theoretical studies on over-smoothing~\cite{oono2019graph,cai2020note,roth2023rank}.
%Ergodic graphs
\begin{corollary}
    Let $\mathbf{A}_1$,\dots,$\mathbf{A}_l$ represent ergodic relations with a weighted mean aggregation. Then, there are no structurally independent nodes.
\end{corollary}
However, graphs without ergodic components behave differently.
Graphs that do not contain any ergodic parts are directed acyclic graphs (DAGs), which we formally define as follows:
\begin{definition}(DAG)
\label{def:dag}
    A graph $G=(\mathcal{V},\mathcal{E})$ is a directed acyclic graph if there exists a strict partial ordering $\prec$ on the nodes such that all edges $(v_i,v_j)\in[n]\times[n]$ satisfy $v_i \prec v_j$. 
\end{definition}

We further refer to nodes without incoming edges as root nodes and to edge relations $\mathbf{E}$ as directed acyclic relations (DARs) if their edges induce a DAG. Utilizing multiple DARs ensures that structurally independent nodes always exist. 

\begin{proposition}
    Let $\mathcal{E}_1, \mathcal{E}_2$ represent two non-empty DARs with different root nodes. Then, there always exist structurally independent nodes for any non-zero edge weights.
\end{proposition}

Thus, constructing multiple DARs is one way to benefit from multiple relations that will result in non-smooth representation even when utilizing the mean as the aggregation function.
Transforming a given edge set into a DAR is a well-known algorithmic challenge in graph theory, which is connected to topological sorting~\citep{cormen2022introduction} and finding feedback arc sets~\citep{garey1979computers}.
In this work, we transform any graph into DARs by defining a strict partial ordering $\prec$ on the nodes. 
The first DAR is obtained by filtering the edges $(i,j)$ to only consider those for which $v_i\prec v_j$, i.e., $\mathcal{E}_1 = \{(v_i,v_j)\in \mathcal{E}\mid v_i \prec v_j\}$. The converse ordering provides a second DAR, i.e., $\mathcal{E}_2 = \{(v_i,v_j)\in \mathcal{E}\mid v_j \prec v_i\}$. 
For some edges, neither $v_i\prec v_j$ nor $v_j\prec v_i$ may be satisfied. These edges are not contained in $\mathcal{E}_1$ or $\mathcal{E}_2$. To retain all edges, we propose a third relation type that contains all the remaining edges, i.e., $\mathcal{E}_3 = \mathcal{E} \setminus (\mathcal{E}_1 \cup \mathcal{E}_2)$. Based on this, we define the graph assignment function to be

\begin{equation}
    f(v_i,v_j) = \begin{cases}
        1, & \text{if } v_i \prec v_j \\
        2, & \text{if } v_j \prec v_i \\
        3, & \text{otherwise}
    \end{cases}   
\end{equation}

We provide an example of such a multi-relational graph in Figure~\ref{fig:comp_graphs}.
We refer to an MRS-MPNN instantiation that uses this relation assignment function as $\textrm{MRS-MPNN}_{\mathrm{DA}}$.
Note that additional graphs cannot reduce the minimal rank of representations. In fact, when $\mathcal{E}_3$ does not have a regularity structure and leads to further structural independent nodes, we can ensure that the minimal rank is at least $3$.

%\subsection{Choosing a partial order}
Many graph theoretical algorithms, such as graph traversal algorithms or centrality measures like the node degree, can provide a strict partial ordering for nodes. %We construct the strict partial order based on a single real node value $r_i$ per node $i$, i.e., $i\prec j \Leftrightarrow r_i < r_j$.
While any partial ordering allows for more informative node representations,
%The choice of partial order adds an inductive bias to the model and removes invariances of messages to their direction, i.e., removing isotropy~\citep{weickert1998anisotropic}.
%For favorable optimization properties, 
the three computational graphs should benefit from different feature transformations. 
Messages within each computational graph should be more similar to each other than to messages of other computational graphs.
For example, when choosing the node degree as our partial ordering, messages sent from higher-degree nodes to lower-degree nodes differ from those sent from lower-degree nodes to higher-degree nodes. The third relation constructs messages between nodes of the same degree. The task should then benefit from different message types between these nodes. In molecular data, higher-degree nodes correspond to other atoms than lower-degree nodes~\citep{wells2012structural}, which may benefit from sending different message types. As this choice has a large effect on the MRS-MPNN, this allows future models to benefit from application-dependent domain knowledge. Many other splitting approaches that do not construct DARs can work similarly well as long as these contain structurally independent nodes.

%\subsection{Properties}
%Assigning each edge to one of $k$ graphs can work, though it needs to be ensured that edge weight sums are different. This is particularly challenging when the mean aggregation is considered. Here, two nodes can only be structurally independent if one node has zero neighbors in a graph, in which the other node does have neighbors.
%\begin{remark}
%Let $v_i,v_j$ be two nodes on graphs $A_p$ and $A_q$. If $s_p[i] = s_q[i]$ and $s_p[j] = s_q[j]$, nodes $v_i$ and $v_j$ cannot be structurally independent. 
%\end{remark}
%\begin{remark}
%    Two nodes $v_i,v_j$ are linearly independent on $\mathbf{A}_p$ and $\mathbf{A}_q$ if $v_i$ does not have neighbors in $\mathbf{A}_p$
%\end{remark}

\begin{table*}[tb]
\caption{Mean values and standard deviations over three runs on the ZINC12k dataset (best results marked in \textbf{bold}). The learning rate and the number of layers are tuned. Train MAE is the overall minimum, while test MAE is based on the best validation MAE. Step times are in milliseconds (ms).}
\label{tab:ordering}
\vskip 0.15in
\begin{center}
\begin{small}
\begin{sc}
\begin{tabular}{lccc}
\toprule
\multirow{2}{*}{Method} & Step Time & \multicolumn{2}{c}{ZINC12k (MAE)} \\
& (ms) & Train & Test \\
\midrule
GCN & $\mathbf{4.3}\pm0.1$ & $0.051\pm0.002$ & $0.404\pm0.011$\\
\midrule
$\textrm{MRS-GCN}_{\textrm{DA}}$ (random) & $5.7\pm0.1$ & $0.006\pm0.001$ & $0.623\pm0.003$\\
$\textrm{MRS-GCN}_{\textrm{DA}}$ (Features) & $5.8\pm0.4$ & $0.011\pm0.003$ & $0.390\pm0.004$\\
$\textrm{MRS-GCN}_{\textrm{DA}}$ (PPR) & $6.8\pm0.4$ & $0.010\pm0.002$ & $0.358\pm0.006$\\
$\textrm{MRS-GCN}_{\textrm{DA}}$ (Degree) & $5.8\pm0.2$ & $\mathbf{0.003}\pm0.001$ & $\mathbf{0.318}\pm0.031$\\
\bottomrule
\end{tabular}
\end{sc}
\end{small}
\end{center}
\vskip -0.1in
\end{table*}

\begin{figure*}[tb]

     \centering
     \begin{minipage}[t]{0.49\textwidth}
         \centering
        \def\svgwidth{\textwidth}
     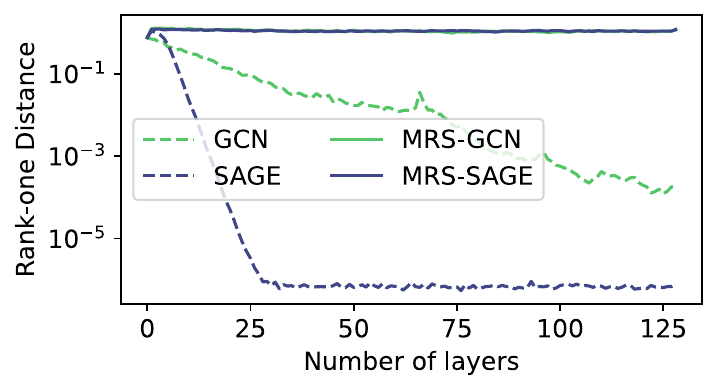
     \caption{Comparison of the Rank-one distance (ROD). Mean values over $50$ random seeds.}
      \label{fig:dirichlet}
     \end{minipage}
     \hfill
     \begin{minipage}[t]{0.49\textwidth}
         \centering
         \def\svgwidth{\textwidth}
         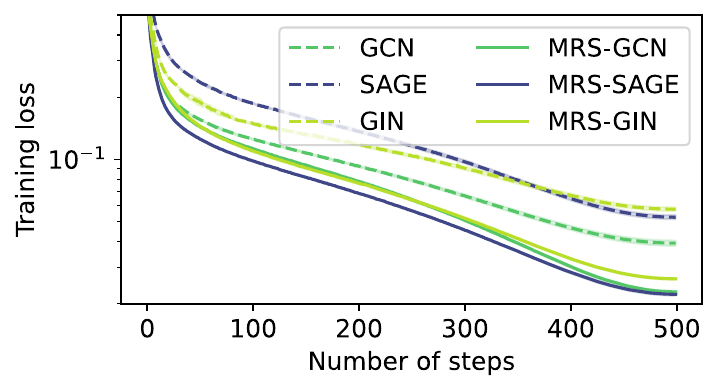
         \caption{Training loss (MAE) during optimization on ZINC. The learning rate and the number of layers are tuned for each MPNN. Mean values over three runs with standard deviations as semi-transparent areas.}
         \label{fig:training_loss}
    \end{minipage}         
\end{figure*}

%\begin{table*}[t]
%\centering
%\caption{Minimum achieved training loss. The cross-entropy (CE) and mean absolute error (MAE) are reported. The learning rate and the number of layers are tuned. Mean and standard deviation over five runs are reported (pairwise best results marked in \textbf{bold}).}
%\label{tab:overview}
%\vskip 0.15in
%\begin{center}
%\begin{small}
%\begin{sc}
%\begin{tabular}{lccc}
%\toprule
%Method & ZINC (MAE) & Peptides-Func (CE) & Peptides-Struct (MAE) \\
%\midrule
%GCN & $0.051\pm0.002$ & $0.033\pm0.008$ & $0.146\pm0.032$\\
%$\textrm{MRS-GCN}_{\textrm{DA}}$ & $\mathbf{0.003}\pm0.001$ & $\mathbf{0.003}\pm0.000$ & $\mathbf{0.052}\pm0.004$ \\
%\midrule
%SAGE & $0.027\pm0.005$ & $0.011\pm0.004$ & $0.168\pm0.004$ \\
%$\textrm{MRS-SAGE}_{\textrm{DA}}$ & $\mathbf{0.002}\pm0.000$ & $\mathbf{0.002}\pm0.000$ & $\mathbf{0.055}\pm0.004$ \\
%\bottomrule
%\end{tabular}
%\end{sc}
%\end{small}
%\end{center}
%\vskip -0.1in
%\end{table*}

\section{Experiments}
\label{sec:exp}
We now investigate the ability of MRS-MPNNs to improve learning for complex tasks. We provide additional details in Appendix~\ref{ap:exp} and reproducible code as supplementary material\footnote{Code is available at \url{https://github.com/roth-andreas/splitting-computational-graphs}}.

\subsection{Improving the Learning Process}
We first consider the challenging ZINC dataset~\citep{sterling2015zinc}. 
%The Peptides dataset consists of \num{15535} graphs. 
%Peptides-Func is a multi-label graph classification task, Peptides-Struct is a multi-label graph regression task, and 
ZINC is a single-label graph regression task. For some of our ablation studies we use the subset containing $\num{12000}$ graphs (ZINC12k), as proposed by \citet{dwivedi2023benchmarking}.
%All considered models perform $k$ iterations of message-passing, with ReLU as a non-linear activation function, followed by a linear prediction layer. 
%We do not use additional techniques like positional encodings, residual connections, normalization layers, or regularization to compare the differences between the message-passing operators independently. 
Following \citet{dwivedi2022LRGB}, all models use at most $\num{500000}$ parameters.
AdamW~\citep{loshchilov2018decoupled} is used for optimization. %See Appendix~\ref{ap:exp} for more details. 
We integrate our models into the implementation of \citet{tonshoff2023did}.

\begin{table*}[tb]
\caption{Mean values and standard deviations over three runs on the ZINC dataset (best results marked in \textbf{bold}). The learning rate and the number of layers are tuned. Train MAE is the overall minimum, while test MAE is based on the best validation MAE. Step times are in milliseconds (ms).}
\label{tab:zinc_full}
\vskip 0.15in
\begin{center}
\begin{small}
\begin{sc}
\begin{tabular}{lcc}
\toprule
\multirow{2}{*}{Method} & \multicolumn{2}{c}{ZINC (MAE)} \\ & Train & Test\\
\midrule
GCN& $0.053\pm0.001$ & $0.155\pm0.003$ \\
$\textrm{MRS-GCN}_{\textrm{DA}}$ & $\mathbf{0.023}\pm0.000$ & $\mathbf{0.134}\pm0.001$ \\
\midrule
SAGE & $0.039\pm0.001$ & $0.123\pm0.002$ \\
$\textrm{MRS-SAGE}_{\textrm{DA}}$ & $\mathbf{0.022}\pm0.000$ & $\mathbf{0.106}\pm0.002$ \\
\midrule
GAT & $0.049\pm0.002$ & $0.149\pm0.001$\\
%GAT+degree & $0.042\pm0.001$ & $0.137\pm0.004$\\
$\textrm{MRS-GAT}_{\textrm{DA}}$ & $\mathbf{0.026}\pm0.001$ & $\mathbf{0.128}\pm0.000$\\
\midrule
GIN & $0.058\pm0.001$ & $0.123\pm0.004$ \\
$\textrm{MRS-GIN}_{\textrm{DA}}$ & $\mathbf{0.026}\pm0.000$ & $\mathbf{0.106}\pm0.004$ \\
\midrule
GatedGCN & $0.051\pm0.002$ & $0.099\pm0.011$\\
$\textrm{MRS-GatedGCN}_{\textrm{DA}}$ & $\mathbf{0.011}\pm0.003$ & $\mathbf{0.088}\pm0.004$\\
\bottomrule
\end{tabular}
\end{sc}
\end{small}
\end{center}
\vskip -0.1in
\end{table*}

%ZINC
\begin{table*}[tb]
\caption{Test results on ZINC12k. The base models are combined with residual connections (Res), jumping knowledge (JK), and Laplacian positional encoding (LapPE). The learning rate is tuned for each entry. Mean average error (MAE) over three runs is reported (pairwise best results marked in \textbf{bold}). Standard deviations are shown in Table~\ref{tab:stdev}.}
\label{tab:layer}
\vskip 0.15in
\begin{center}
\footnotesize
\begin{tabular}{lcccccc}
\toprule
Method & 1 & 2 & 4 & 8 & 16 & 32 \\
\midrule
GCN & $0.591$ & $0.493$ & $0.421$ & $0.404$ & $0.417$ & $0.440$ \\
$\textrm{MRS-GCN}_{\textrm{DA}}$ & $\mathbf{0.525}$ & $\mathbf{0.445}$ & $\mathbf{0.343}$ & $\mathbf{0.318}$ & $\mathbf{0.318}$ & $\mathbf{0.338}$ \\
\midrule
GCN + Res & $0.567$ & $0.471$ & $0.427$ & $0.403$ & $0.370$ & $0.336$ \\
$\textrm{MRS-GCN}_{\textrm{DA}}$ + Res & $\mathbf{0.508}$ & $\mathbf{0.402}$ & $\mathbf{0.295}$ & $\mathbf{0.257}$ & $\mathbf{0.252}$ & $\mathbf{0.250}$\\
\midrule
GCN + JK & $0.588$ & $0.496$ & $0.424$ & $0.409$ & $0.413$ & $0.435$\\
$\textrm{MRS-GCN}_{\textrm{DA}}$ + JK & $\mathbf{0.526}$ & $\mathbf{0.442}$ & $\mathbf{0.324}$ & $\mathbf{0.303}$ & $\mathbf{0.311}$ & $\mathbf{0.304}$ \\
\midrule
GCN + LapPE & $0.498$ & $0.437$ & $0.392$ & $0.367$ & $0.383$ & $0.444$ \\
$\textrm{MRS-GCN}_{\textrm{DA}}$ + LapPE & $\mathbf{0.441}$ & $\mathbf{0.363}$ & $\mathbf{0.292}$ & $\mathbf{0.272}$ & $\mathbf{0.297}$ & $\mathbf{0.317}$ \\
\bottomrule
\end{tabular}
\end{center}
\vskip -0.1in
\end{table*}

\paragraph{Evaluating Node Orderings}
We evaluate several strategies for constructing node orderings using the ZINC12k dataset. We consider random values, the sum of initial node features, Personalized PageRank (PPR) scores, and the node degree. The strategies are evaluated in terms of execution time per training step, minimal achieved training loss, and test loss corresponding to the best validation performance. We tune the learning rate and the number of layers for each method. All orderings are applied to $\textrm{MRS-GCN}_\textrm{DA}$.

Results are presented in Table~\ref{tab:ordering}. While the GCN underfits the training data, all MRS-GCNs significantly improve the training loss. The random ordering worsens the test performance. This highlights the importance of constructing relations that align with the given data and task.
For the other orderings, the test performance is improved as a consequence of the improved training loss.
The degree-based ordering achieves the best train and test performance. Runtime is increased by around $35\%$ using our implementation. A detailed runtime evaluation is provided in Table~\ref{tab:runtime}. The notable differences in performance between orderings show potential for optimal task-dependent orderings and graph splittings. To study further properties of multi-relational MPNNs, we utilize the node degree as our ordering for all other experiments.

\paragraph{More Informative Node Representations}
To empirically validate that the representations obtained by MRS-MPNNs can be more informative than those obtained by MPNNs, we compare their distance to a rank-one matrix throughout $128$ iterations using the rank-one distance (ROD)~\citep{roth2024simplifying}. In each iteration, one message-passing step and the ReLU activation function are applied, and ROD is calculated. We repeat this process for $50$ random graphs from the ZINC dataset.
Results for GCN, $\textrm{MRS-GCN}_{\textrm{DA}}$, SAGE, and $\textrm{MRS-SAGE}_{\textrm{DA}}$ are shown in Figure~\ref{fig:dirichlet}. 
The representations for GCN and SAGE converge to a rank-one state, while the distance to a rank-one matrix remains roughly constant across all $128$ layers for the DA versions.

\paragraph{Additional Experiments}
To better understand the advantages of MRS-MPNNs, we present training and test performances for ZINC using the GCN, SAGE, GAT~\citep{velivckovic2018graph}, GIN~\citep{xu2018how}, GatedGCN~\citep{dwivedi2023benchmarking}, and their corresponding MRS-MPNNs in Table~\ref{tab:zinc_full}. We provide details about each model in Appendix~\ref{sec:model_details}. In all cases, the respective MRS-MPNNs achieve a significantly reduced training loss. Due to the ability of MRS-MPNNs to amplify multiple signals, they suffer much less from underfitting and consequently improve the test performance. To emphasize this, we display the training loss during optimization in Figure~\ref{fig:training_loss}. MRS-MPNNs visibly improve the optimization process. 

GCNs are often combined with more advanced techniques, like residual connections (Res)~\citep{he2016deep}, jumping knowledge (JK)~\citep{xu2018representation}, and Laplacian positional encodings~\citep{kreuzer2021rethinking}. We evaluate their interplay with MRS-MPNNs for various layers and present test scores in Table~\ref{tab:layer}. MRS-GCNs outperform their corresponding GCNs in all cases, and achieve their best performance with deeper models. We note a slight drop in performance for $32$ layers in some cases for MRS-GCNs. This can be attributed to optimization challenges, i.e., exploding and vanishing gradients, as normalization layers were not considered here.

\subsection{Comparison with State-of-the-Art}
Improved optimization of the target function suggests that MRS-MPNNs will be particularly valuable for challenging tasks. We consider large-scale heterophilic graphs, namely Squirrel and Chameleon~\citep{Pei2020Geom-GCN}, Arxiv-Year and Snap-Patents~\citep{lim2021large}, and Roman-Empire~\citep{platonov2023a}.
Our implementation is based on Dir-GNN~\citep{rossi2023edge}, a state-of-the-art method. We replace the MPNN modules of Dir-GNN with the corresponding $\textrm{MRS-MPNN}_{\textrm{DA}}$ modules. Accordingly, $\textrm{MRS-SAGE}_{\textrm{DA}}$ is used for Roman-Empire, while $\textrm{MRS-GCN}_{\textrm{DA}}$ is used for the other datasets. Experiments for Squirrel, Chameleon, and Roman-Empire are repeated for ten fixed splits into training, validation, and test sets and for Arxiv-Year and Snap-Patents for five fixed splits. 
The baseline results are reused from our reference implementation by~\citet{rossi2023edge}.
Based on their hyperparameter values, we tune the learning rate, number of layers, and dropout ratio using a grid search. %Best-performing hyperparameters are reused for five repetitions of each split, for which we report the average test scores. 
We compare to seven state-of-the-art methods for heterophilic graphs and three for directed graphs. Further experimental details are provided in Appendix~\ref{ap:exp}. 

We present the test results in Table~\ref{tab:hetero}. MRS-MPNNs slightly improve the performance for all five datasets, with more significant gains for the larger datasets, i.e., Arxiv-Year and Snap-Patents. As with our other experiments, we observe larger improvements in the training loss. %This confirms the benefits of MRS-MPNNs for optimization, while generalization properties can be further improved.

\begin{table*}[tb]
\caption{Mean accuracy and standard deviation for five directed benchmark graphs (best result marked in \textbf{bold} and second-best \underline{underlined}).}
\label{tab:hetero}
\vskip 0.15in
\begin{center}
\begin{small}
\begin{sc}
\begin{tabular}{lccccc}
\toprule
Method & Squirrel & Chameleon & Arxiv-year & Snap-patents & Roman-Empire \\
\midrule
MLP    & $28.77\pm1.56$ & $46.21\pm2.99$ & $36.70\pm0.21$ & $31.34\pm0.05$ & $64.94\pm0.62$ \\
GCN & $53.43\pm2.01$ & $64.82\pm2.24$ & $46.02\pm0.26$ & $51.02\pm0.06$ & $73.69\pm0.74$ \\
H\_2GCN & $37.90\pm2.02$ & $59.39\pm1.98$ & $49.09\pm0.10$ & OOM & $60.11\pm0.52$ \\
GPR-GNN & $54.35\pm0.87$ & $62.85\pm2.90$ & $45.07\pm0.21$ & $40.19\pm0.03$ & $64.85\pm0.27$ \\
LINKX & $61.81\pm1.80$ & $68.42\pm1.38$ & $56.00\pm0.17$ & $61.95\pm0.12$ & $37.55\pm0.36$ \\
FSGNN & $74.10\pm1.89$ & $78.27\pm1.28$ & $50.47\pm0.21$ & $65.07\pm0.03$ & $79.92\pm0.56$ \\
ACM-GCN & $67.40\pm2.21$ & $74.76\pm2.20$ & $47.37\pm0.59$ & $55.14\pm0.16$ & $69.66\pm0.62$ \\
GloGNN & $57.88\pm1.76$ & $71.21\pm1.84$ & $54.79\pm0.25$ & $62.09\pm0.27$ & $59.63\pm0.69$ \\
Grad. Gating & $64.26\pm2.38$ & $71.40\pm2.38$ & $63.30\pm1.84$ & $69.50\pm0.39$ & $82.16\pm0.78$ \\
\midrule
DiGCN & $37.74\pm1.54$ & $52.24\pm3.65$ & OOM & OOM & $52.71\pm0.32$ \\
MagNet & $39.01\pm1.93$ & $58.22\pm2.87$ & $60.29\pm0.27$ & OOM & $88.07\pm0.27$ \\
Dir-GNN & $\underline{75.31}\pm1.92$ & $\underline{79.71}\pm1.26$ & $\underline{64.08}\pm0.30$ & $\underline{73.95}\pm0.05$ & $\underline{91.23}\pm0.32$ \\
%FaberNet & $\mathbf{76.71}\pm1.92$ & $\mathbf{80.33}\pm1.19$ & $\underline{64.62}\pm1.01$ & $\underline{75.10}\pm0.03$ & $\underline{92.24}\pm0.43$ \\
\midrule
$\textrm{MRS-Dir-GCN}_{\textrm{DA}}$ & $\mathbf{76.01}\pm1.90$ & $\mathbf{80.17}\pm1.88$ & $\mathbf{66.03}\pm0.20$ & $\mathbf{74.72}\pm0.05$ & $\mathbf{91.87}\pm0.42$ \\
\bottomrule
\end{tabular}
\end{sc}
\end{small}
\end{center}
\vskip -0.1in
\end{table*}
\section{Conclusion}
\label{sec:conclusion}
% Summary
In this work, we propose to split graphs into multi-relational graphs and operate MPNNs on these. We identify the necessary and sufficient condition on these relations that ensures that representations will always have more linearly independent features. While we show that this is always satisfied when operating on multiple DARs, many other graph splitting techniques can be designed based on our theory. Our experiments confirm that operating with multiple relations results in more informative node representations, which improves learning.
% Limitations
As limitations of MR-MPNNs, we have seen an increase in runtime, the need to find a suitable graph splitting method and that they tend to overfit on the training data.
% Future Work
We anticipate further opportunities for other graph splitting methods that satisfy our condition. Task-specific knowledge and invariances for graph splitting can provide additional benefits. 
% Limitations
%Limitations of this work may be the 

%\section*{Author Contributions}
%Authors of accepted papers are \emph{encouraged} to include a statement that declares the individual contribution of every author, especially when there are co-authors that made equal contributions to the research.
%You may adopt the \href{https://credit.niso.org/}{Contributor Roles Taxonomy (CRediT)} methodology for attributing contributions.
%Do not include this section in the version for blind review.
%This section does not count towards the page limit.

\section*{Acknowledgements}
Part of this research has been funded by the Federal Ministry of Education and Research of Germany and the state of North-Rhine Westphalia as part of the Lamarr-Institute for Machine Learning and Artificial Intelligence and by the Federal Ministry of Education and Research of Germany under grant no. 01IS22094E WEST-AI. This work was supported by the Vienna Science and
Technology Fund (WWTF) [10.47379/VRG19009]. Simulations were performed with computing resources granted by WestAI under project rwth1631.

%The \LaTeX{} template of LoG 2024 is heavily borrowed from the NeurIPS template.

%Do not include acknowledgements in the version for blind review.
%If a paper is accepted, please place such acknowledgements in an unnumbered section at the end of the paper, immediately before the references.
%The acknowledgements do not count towards the page limit.

% For natbib users:
\bibliographystyle{unsrtnat}
\bibliography{references}

\begin{thebibliography}{78}
\providecommand{\natexlab}[1]{#1}
\providecommand{\url}[1]{\texttt{#1}}
\expandafter\ifx\csname urlstyle\endcsname\relax
  \providecommand{\doi}[1]{doi: #1}\else
  \providecommand{\doi}{doi: \begingroup \urlstyle{rm}\Url}\fi

\bibitem[Igashov et~al.(2024)Igashov, St{\"{a}}rk, Vignac, Schneuing, Satorras, Frossard, Welling, Bronstein, and Correia]{igashov2024equivariant}
Ilia Igashov, Hannes St{\"{a}}rk, Cl{\'{e}}ment Vignac, Arne Schneuing, Victor~Garcia Satorras, Pascal Frossard, Max Welling, Michael~M. Bronstein, and Bruno~E. Correia.
\newblock Equivariant 3d-conditional diffusion model for molecular linker design.
\newblock \emph{Nat. Mac. Intell.}, 6\penalty0 (4):\penalty0 417--427, 2024.
\newblock \doi{10.1038/S42256-024-00815-9}.

\bibitem[Fan et~al.(2019)Fan, Ma, Li, He, Zhao, Tang, and Yin]{fan2019graph}
Wenqi Fan, Yao Ma, Qing Li, Yuan He, Yihong~Eric Zhao, Jiliang Tang, and Dawei Yin.
\newblock Graph neural networks for social recommendation.
\newblock In Ling Liu, Ryen~W. White, Amin Mantrach, Fabrizio Silvestri, Julian~J. McAuley, Ricardo Baeza{-}Yates, and Leila Zia, editors, \emph{The World Wide Web Conference, {WWW} 2019, San Francisco, CA, USA, May 13-17, 2019}, pages 417--426. {ACM}, 2019.
\newblock \doi{10.1145/3308558.3313488}.

\bibitem[Derrow{-}Pinion et~al.(2021)Derrow{-}Pinion, She, Wong, Lange, Hester, Perez, Nunkesser, Lee, Guo, Wiltshire, Battaglia, Gupta, Li, Xu, Sanchez{-}Gonzalez, Li, and Velickovic]{derrow2021eta}
Austin Derrow{-}Pinion, Jennifer She, David Wong, Oliver Lange, Todd Hester, Luis Perez, Marc Nunkesser, Seongjae Lee, Xueying Guo, Brett Wiltshire, Peter~W. Battaglia, Vishal Gupta, Ang Li, Zhongwen Xu, Alvaro Sanchez{-}Gonzalez, Yujia Li, and Petar Velickovic.
\newblock {ETA} prediction with graph neural networks in google maps.
\newblock In Gianluca Demartini, Guido Zuccon, J.~Shane Culpepper, Zi~Huang, and Hanghang Tong, editors, \emph{{CIKM} '21: The 30th {ACM} International Conference on Information and Knowledge Management, Virtual Event, Queensland, Australia, November 1 - 5, 2021}, pages 3767--3776. {ACM}, 2021.
\newblock \doi{10.1145/3459637.3481916}.

\bibitem[Gilmer et~al.(2017)Gilmer, Schoenholz, Riley, Vinyals, and Dahl]{gilmer2017neural}
Justin Gilmer, Samuel~S. Schoenholz, Patrick~F. Riley, Oriol Vinyals, and George~E. Dahl.
\newblock Neural message passing for quantum chemistry.
\newblock In Doina Precup and Yee~Whye Teh, editors, \emph{Proceedings of the 34th International Conference on Machine Learning, {ICML} 2017, Sydney, NSW, Australia, 6-11 August 2017}, volume~70 of \emph{Proceedings of Machine Learning Research}, pages 1263--1272. {PMLR}, 2017.

\bibitem[Kriege(2022)]{NEURIPS2022_7eed2822}
Nils~M. Kriege.
\newblock Weisfeiler and leman go walking: Random walk kernels revisited.
\newblock In Sanmi Koyejo, S.~Mohamed, A.~Agarwal, Danielle Belgrave, K.~Cho, and A.~Oh, editors, \emph{Advances in Neural Information Processing Systems 35: Annual Conference on Neural Information Processing Systems 2022, NeurIPS 2022, New Orleans, LA, USA, November 28 - December 9, 2022}, 2022.

\bibitem[He et~al.(2016)He, Zhang, Ren, and Sun]{he2016deep}
Kaiming He, Xiangyu Zhang, Shaoqing Ren, and Jian Sun.
\newblock Deep residual learning for image recognition.
\newblock In \emph{2016 {IEEE} Conference on Computer Vision and Pattern Recognition, {CVPR} 2016, Las Vegas, NV, USA, June 27-30, 2016}, pages 770--778. {IEEE} Computer Society, 2016.
\newblock \doi{10.1109/CVPR.2016.90}.

\bibitem[Vaswani et~al.(2017)Vaswani, Shazeer, Parmar, Uszkoreit, Jones, Gomez, Kaiser, and Polosukhin]{vaswani2017attention}
Ashish Vaswani, Noam Shazeer, Niki Parmar, Jakob Uszkoreit, Llion Jones, Aidan~N. Gomez, Lukasz Kaiser, and Illia Polosukhin.
\newblock Attention is all you need.
\newblock In Isabelle Guyon, Ulrike von Luxburg, Samy Bengio, Hanna~M. Wallach, Rob Fergus, S.~V.~N. Vishwanathan, and Roman Garnett, editors, \emph{Advances in Neural Information Processing Systems 30: Annual Conference on Neural Information Processing Systems 2017, December 4-9, 2017, Long Beach, CA, {USA}}, pages 5998--6008, 2017.

\bibitem[Li et~al.(2018)Li, Han, and Wu]{li2018deeper}
Qimai Li, Zhichao Han, and Xiao{-}Ming Wu.
\newblock Deeper insights into graph convolutional networks for semi-supervised learning.
\newblock In Sheila~A. McIlraith and Kilian~Q. Weinberger, editors, \emph{Proceedings of the Thirty-Second {AAAI} Conference on Artificial Intelligence, (AAAI-18), the 30th innovative Applications of Artificial Intelligence (IAAI-18), and the 8th {AAAI} Symposium on Educational Advances in Artificial Intelligence (EAAI-18), New Orleans, Louisiana, USA, February 2-7, 2018}, pages 3538--3545. {AAAI} Press, 2018.
\newblock \doi{10.1609/AAAI.V32I1.11604}.

\bibitem[Yan et~al.(2022)Yan, Hashemi, Swersky, Yang, and Koutra]{yan2022two}
Yujun Yan, Milad Hashemi, Kevin Swersky, Yaoqing Yang, and Danai Koutra.
\newblock Two sides of the same coin: Heterophily and oversmoothing in graph convolutional neural networks.
\newblock In Xingquan Zhu, Sanjay Ranka, My~T. Thai, Takashi Washio, and Xindong Wu, editors, \emph{{IEEE} International Conference on Data Mining, {ICDM} 2022, Orlando, FL, USA, November 28 - Dec. 1, 2022}, pages 1287--1292. {IEEE}, 2022.
\newblock \doi{10.1109/ICDM54844.2022.00169}.

\bibitem[Roth and Liebig(2023)]{roth2023rank}
Andreas Roth and Thomas Liebig.
\newblock Rank collapse causes over-smoothing and over-correlation in graph neural networks.
\newblock In Soledad Villar and Benjamin Chamberlain, editors, \emph{Learning on Graphs Conference, 27-30 November 2023, Virtual Event}, volume 231 of \emph{Proceedings of Machine Learning Research}, page~35. {PMLR}, 2023.

\bibitem[Roth(2024)]{roth2024simplifying}
Andreas Roth.
\newblock Simplifying the theory on over-smoothing.
\newblock \emph{CoRR}, abs/2407.11876, 2024.
\newblock \doi{10.48550/ARXIV.2407.11876}.

\bibitem[Chen et~al.(2020)Chen, Wei, Huang, Ding, and Li]{chen2020simple}
Ming Chen, Zhewei Wei, Zengfeng Huang, Bolin Ding, and Yaliang Li.
\newblock Simple and deep graph convolutional networks.
\newblock In \emph{Proceedings of the 37th International Conference on Machine Learning, {ICML} 2020, 13-18 July 2020, Virtual Event}, volume 119 of \emph{Proceedings of Machine Learning Research}, pages 1725--1735. {PMLR}, 2020.

\bibitem[Scholkemper et~al.(2024)Scholkemper, Wu, Jadbabaie, and Schaub]{scholkemper2024residual}
Michael Scholkemper, Xinyi Wu, Ali Jadbabaie, and Michael~T. Schaub.
\newblock Residual connections and normalization can provably prevent oversmoothing in gnns.
\newblock \emph{CoRR}, abs/2406.02997, 2024.
\newblock \doi{10.48550/ARXIV.2406.02997}.

\bibitem[Roth and Liebig(2022)]{roth2022transforming}
Andreas Roth and Thomas Liebig.
\newblock Transforming pagerank into an infinite-depth graph neural network.
\newblock In Massih{-}Reza Amini, St{\'{e}}phane Canu, Asja Fischer, Tias Guns, Petra~Kralj Novak, and Grigorios Tsoumakas, editors, \emph{Machine Learning and Knowledge Discovery in Databases - European Conference, {ECML} {PKDD} 2022, Grenoble, France, September 19-23, 2022, Proceedings, Part {II}}, volume 13714 of \emph{Lecture Notes in Computer Science}, pages 469--484. Springer, 2022.
\newblock \doi{10.1007/978-3-031-26390-3\_27}.

\bibitem[Li et~al.(2016)Li, Tarlow, Brockschmidt, and Zemel]{li2015gated}
Yujia Li, Daniel Tarlow, Marc Brockschmidt, and Richard~S. Zemel.
\newblock Gated graph sequence neural networks.
\newblock In Yoshua Bengio and Yann LeCun, editors, \emph{4th International Conference on Learning Representations, {ICLR} 2016, San Juan, Puerto Rico, May 2-4, 2016, Conference Track Proceedings}, 2016.

\bibitem[Rusch et~al.(2023)Rusch, Chamberlain, Mahoney, Bronstein, and Mishra]{rusch2022gradient}
T.~Konstantin Rusch, Benjamin~Paul Chamberlain, Michael~W. Mahoney, Michael~M. Bronstein, and Siddhartha Mishra.
\newblock Gradient gating for deep multi-rate learning on graphs.
\newblock In \emph{The Eleventh International Conference on Learning Representations, {ICLR} 2023, Kigali, Rwanda, May 1-5, 2023}. OpenReview.net, 2023.

\bibitem[Oono and Suzuki(2020)]{oono2019graph}
Kenta Oono and Taiji Suzuki.
\newblock Graph neural networks exponentially lose expressive power for node classification.
\newblock In \emph{8th International Conference on Learning Representations, {ICLR} 2020, Addis Ababa, Ethiopia, April 26-30, 2020}. OpenReview.net, 2020.

\bibitem[Cai and Wang(2020)]{cai2020note}
Chen Cai and Yusu Wang.
\newblock A note on over-smoothing for graph neural networks.
\newblock \emph{CoRR}, abs/2006.13318, 2020.

\bibitem[Giovanni et~al.(2023)Giovanni, Rowbottom, Chamberlain, Markovich, and Bronstein]{digiovanni2023understanding}
Francesco~Di Giovanni, James Rowbottom, Benjamin~Paul Chamberlain, Thomas Markovich, and Michael~M. Bronstein.
\newblock Understanding convolution on graphs via energies.
\newblock \emph{Trans. Mach. Learn. Res.}, 2023, 2023.

\bibitem[Bresson and Laurent(2017)]{bresson2017residual}
Xavier Bresson and Thomas Laurent.
\newblock Residual gated graph convnets.
\newblock \emph{CoRR}, abs/1711.07553, 2017.

\bibitem[Klicpera et~al.(2018)Klicpera, Bojchevski, and G{\"{u}}nnemann]{gasteiger2018combining}
Johannes Klicpera, Aleksandar Bojchevski, and Stephan G{\"{u}}nnemann.
\newblock Personalized embedding propagation: Combining neural networks on graphs with personalized pagerank.
\newblock \emph{CoRR}, abs/1810.05997, 2018.

\bibitem[Finkelshtein et~al.(2024)Finkelshtein, Huang, Bronstein, and Ceylan]{finkelshtein2023cooperative}
Ben Finkelshtein, Xingyue Huang, Michael~M. Bronstein, and {\.I}smail~{\.I}lkan Ceylan.
\newblock Cooperative graph neural networks.
\newblock In \emph{Forty-first International Conference on Machine Learning, {ICML} 2024, Vienna, Austria, July 21-27, 2024}. OpenReview.net, 2024.

\bibitem[Zhao and Akoglu(2020)]{Zhao2020PairNorm}
Lingxiao Zhao and Leman Akoglu.
\newblock Pairnorm: Tackling oversmoothing in gnns.
\newblock In \emph{8th International Conference on Learning Representations, {ICLR} 2020, Addis Ababa, Ethiopia, April 26-30, 2020}. OpenReview.net, 2020.

\bibitem[Li et~al.(2023)Li, Xiong, Qian, Thabet, and Ghanem]{li2020deepergcn}
Guohao Li, Chenxin Xiong, Guocheng Qian, Ali~K. Thabet, and Bernard Ghanem.
\newblock Deepergcn: Training deeper gcns with generalized aggregation functions.
\newblock \emph{{IEEE} Trans. Pattern Anal. Mach. Intell.}, 45\penalty0 (11):\penalty0 13024--13034, 2023.
\newblock \doi{10.1109/TPAMI.2023.3306930}.

\bibitem[Zhou et~al.(2021)Zhou, Huang, Zha, Chen, Li, Choi, and Hu]{zhou2021dirichlet}
Kaixiong Zhou, Xiao Huang, Daochen Zha, Rui Chen, Li~Li, Soo{-}Hyun Choi, and Xia Hu.
\newblock Dirichlet energy constrained learning for deep graph neural networks.
\newblock In Marc'Aurelio Ranzato, Alina Beygelzimer, Yann~N. Dauphin, Percy Liang, and Jennifer~Wortman Vaughan, editors, \emph{Advances in Neural Information Processing Systems 34: Annual Conference on Neural Information Processing Systems 2021, NeurIPS 2021, December 6-14, 2021, virtual}, pages 21834--21846, 2021.

\bibitem[Eliasof et~al.(2023{\natexlab{a}})Eliasof, Ruthotto, and Treister]{eliasof2023improving}
Moshe Eliasof, Lars Ruthotto, and Eran Treister.
\newblock Improving graph neural networks with learnable propagation operators.
\newblock In Andreas Krause, Emma Brunskill, Kyunghyun Cho, Barbara Engelhardt, Sivan Sabato, and Jonathan Scarlett, editors, \emph{International Conference on Machine Learning, {ICML} 2023, 23-29 July 2023, Honolulu, Hawaii, {USA}}, volume 202 of \emph{Proceedings of Machine Learning Research}, pages 9224--9245. {PMLR}, 2023{\natexlab{a}}.

\bibitem[Bo et~al.(2021)Bo, Wang, Shi, and Shen]{bo2021beyond}
Deyu Bo, Xiao Wang, Chuan Shi, and Huawei Shen.
\newblock Beyond low-frequency information in graph convolutional networks.
\newblock In \emph{Thirty-Fifth {AAAI} Conference on Artificial Intelligence, {AAAI} 2021, Thirty-Third Conference on Innovative Applications of Artificial Intelligence, {IAAI} 2021, The Eleventh Symposium on Educational Advances in Artificial Intelligence, {EAAI} 2021, Virtual Event, February 2-9, 2021}, pages 3950--3957. {AAAI} Press, 2021.
\newblock \doi{10.1609/AAAI.V35I5.16514}.

\bibitem[Corso et~al.(2020)Corso, Cavalleri, Beaini, Li{\`{o}}, and Velickovic]{corso2020principal}
Gabriele Corso, Luca Cavalleri, Dominique Beaini, Pietro Li{\`{o}}, and Petar Velickovic.
\newblock Principal neighbourhood aggregation for graph nets.
\newblock In Hugo Larochelle, Marc'Aurelio Ranzato, Raia Hadsell, Maria{-}Florina Balcan, and Hsuan{-}Tien Lin, editors, \emph{Advances in Neural Information Processing Systems 33: Annual Conference on Neural Information Processing Systems 2020, NeurIPS 2020, December 6-12, 2020, virtual}, 2020.

\bibitem[Tailor et~al.(2022)Tailor, Opolka, Li{\`{o}}, and Lane]{tailor2022adaptive}
Shyam~A. Tailor, Felix~L. Opolka, Pietro Li{\`{o}}, and Nicholas~Donald Lane.
\newblock Do we need anisotropic graph neural networks?
\newblock In \emph{The Tenth International Conference on Learning Representations, {ICLR} 2022, Virtual Event, April 25-29, 2022}. OpenReview.net, 2022.

\bibitem[Jin et~al.(2022)Jin, Liu, Ma, Aggarwal, and Tang]{jin2022feature}
Wei Jin, Xiaorui Liu, Yao Ma, Charu~C. Aggarwal, and Jiliang Tang.
\newblock Feature overcorrelation in deep graph neural networks: {A} new perspective.
\newblock In Aidong Zhang and Huzefa Rangwala, editors, \emph{{KDD} '22: The 28th {ACM} {SIGKDD} Conference on Knowledge Discovery and Data Mining, Washington, DC, USA, August 14 - 18, 2022}, pages 709--719. {ACM}, 2022.
\newblock \doi{10.1145/3534678.3539445}.

\bibitem[Alon and Yahav(2021)]{alon2021on}
Uri Alon and Eran Yahav.
\newblock On the bottleneck of graph neural networks and its practical implications.
\newblock In \emph{9th International Conference on Learning Representations, {ICLR} 2021, Virtual Event, Austria, May 3-7, 2021}. OpenReview.net, 2021.

\bibitem[Topping et~al.(2022)Topping, Giovanni, Chamberlain, Dong, and Bronstein]{topping2022understanding}
Jake Topping, Francesco~Di Giovanni, Benjamin~Paul Chamberlain, Xiaowen Dong, and Michael~M. Bronstein.
\newblock Understanding over-squashing and bottlenecks on graphs via curvature.
\newblock In \emph{The Tenth International Conference on Learning Representations, {ICLR} 2022, Virtual Event, April 25-29, 2022}. OpenReview.net, 2022.

\bibitem[Abboud et~al.(2022)Abboud, Dimitrov, and Ceylan]{abboud2022shortest}
Ralph Abboud, Radoslav Dimitrov, and {\.I}smail~{\.I}lkan Ceylan.
\newblock Shortest path networks for graph property prediction.
\newblock In Bastian Rieck and Razvan Pascanu, editors, \emph{Learning on Graphs Conference, LoG 2022, 9-12 December 2022, Virtual Event}, volume 198 of \emph{Proceedings of Machine Learning Research}, page~5. {PMLR}, 2022.

\bibitem[Barbero et~al.(2024)Barbero, Velingker, Saberi, Bronstein, and Giovanni]{barbero2024localityaware}
Federico Barbero, Ameya Velingker, Amin Saberi, Michael~M. Bronstein, and Francesco~Di Giovanni.
\newblock Locality-aware graph rewiring in gnns.
\newblock In \emph{The Twelfth International Conference on Learning Representations, {ICLR} 2024, Vienna, Austria, May 7-11, 2024}. OpenReview.net, 2024.

\bibitem[Nguyen et~al.(2023)Nguyen, Hieu, Nguyen, Ho, Osher, and Nguyen]{pmlr-v202-nguyen23c}
Khang Nguyen, Nong~Minh Hieu, Vinh~Duc Nguyen, Nhat Ho, Stanley~J. Osher, and Tan~Minh Nguyen.
\newblock Revisiting over-smoothing and over-squashing using ollivier-ricci curvature.
\newblock In Andreas Krause, Emma Brunskill, Kyunghyun Cho, Barbara Engelhardt, Sivan Sabato, and Jonathan Scarlett, editors, \emph{International Conference on Machine Learning, {ICML} 2023, 23-29 July 2023, Honolulu, Hawaii, {USA}}, volume 202 of \emph{Proceedings of Machine Learning Research}, pages 25956--25979. {PMLR}, 2023.

\bibitem[Schlichtkrull et~al.(2018)Schlichtkrull, Kipf, Bloem, van~den Berg, Titov, and Welling]{schlichtkrull2018modeling}
Michael~Sejr Schlichtkrull, Thomas~N. Kipf, Peter Bloem, Rianne van~den Berg, Ivan Titov, and Max Welling.
\newblock Modeling relational data with graph convolutional networks.
\newblock In Aldo Gangemi, Roberto Navigli, Maria{-}Esther Vidal, Pascal Hitzler, Rapha{\"{e}}l Troncy, Laura Hollink, Anna Tordai, and Mehwish Alam, editors, \emph{The Semantic Web - 15th International Conference, {ESWC} 2018, Heraklion, Crete, Greece, June 3-7, 2018, Proceedings}, volume 10843 of \emph{Lecture Notes in Computer Science}, pages 593--607. Springer, 2018.
\newblock \doi{10.1007/978-3-319-93417-4\_38}.

\bibitem[Vashishth et~al.(2020)Vashishth, Sanyal, Nitin, and Talukdar]{Vashishth2020Composition}
Shikhar Vashishth, Soumya Sanyal, Vikram Nitin, and Partha~P. Talukdar.
\newblock Composition-based multi-relational graph convolutional networks.
\newblock In \emph{8th International Conference on Learning Representations, {ICLR} 2020, Addis Ababa, Ethiopia, April 26-30, 2020}. OpenReview.net, 2020.

\bibitem[Suresh et~al.(2021)Suresh, Budde, Neville, Li, and Ma]{suresh2021breaking}
Susheel Suresh, Vinith Budde, Jennifer Neville, Pan Li, and Jianzhu Ma.
\newblock Breaking the limit of graph neural networks by improving the assortativity of graphs with local mixing patterns.
\newblock In Feida Zhu, Beng~Chin Ooi, and Chunyan Miao, editors, \emph{{KDD} '21: The 27th {ACM} {SIGKDD} Conference on Knowledge Discovery and Data Mining, Virtual Event, Singapore, August 14-18, 2021}, pages 1541--1551. {ACM}, 2021.
\newblock \doi{10.1145/3447548.3467373}.

\bibitem[Yang et~al.(2020)Yang, Feng, Song, and Wang]{yang2020factorizable}
Yiding Yang, Zunlei Feng, Mingli Song, and Xinchao Wang.
\newblock Factorizable graph convolutional networks.
\newblock In Hugo Larochelle, Marc'Aurelio Ranzato, Raia Hadsell, Maria{-}Florina Balcan, and Hsuan{-}Tien Lin, editors, \emph{Advances in Neural Information Processing Systems 33: Annual Conference on Neural Information Processing Systems 2020, NeurIPS 2020, December 6-12, 2020, virtual}, 2020.

\bibitem[Guo et~al.(2022)Guo, Huang, Yi, and Zhang]{guo2022esgnn}
Jingwei Guo, Kaizhu Huang, Xinping Yi, and Rui Zhang.
\newblock {ES-GNN:} generalizing graph neural networks beyond homophily with edge splitting.
\newblock \emph{CoRR}, abs/2205.13700, 2022.
\newblock \doi{10.48550/ARXIV.2205.13700}.

\bibitem[Giunchiglia et~al.(2022)Giunchiglia, Shukla, Gonzalez, and Agarwal]{giunchiglia2022towards}
Valentina Giunchiglia, Chirag~Varun Shukla, Guadalupe Gonzalez, and Chirag Agarwal.
\newblock Towards training gnns using explanation directed message passing.
\newblock In Bastian Rieck and Razvan Pascanu, editors, \emph{Learning on Graphs Conference, LoG 2022, 9-12 December 2022, Virtual Event}, volume 198 of \emph{Proceedings of Machine Learning Research}, page~28. {PMLR}, 2022.

\bibitem[Luan et~al.(2022)Luan, Hua, Lu, Zhu, Zhao, Zhang, Chang, and Precup]{luan2022revisiting}
Sitao Luan, Chenqing Hua, Qincheng Lu, Jiaqi Zhu, Mingde Zhao, Shuyuan Zhang, Xiao{-}Wen Chang, and Doina Precup.
\newblock Revisiting heterophily for graph neural networks.
\newblock In Sanmi Koyejo, S.~Mohamed, A.~Agarwal, Danielle Belgrave, K.~Cho, and A.~Oh, editors, \emph{Advances in Neural Information Processing Systems 35: Annual Conference on Neural Information Processing Systems 2022, NeurIPS 2022, New Orleans, LA, USA, November 28 - December 9, 2022}, 2022.

\bibitem[Eliasof et~al.(2023{\natexlab{b}})Eliasof, Haber, and Treister]{eliasof2023adrgnn}
Moshe Eliasof, Eldad Haber, and Eran Treister.
\newblock {ADR-GNN:} advection-diffusion-reaction graph neural networks.
\newblock \emph{CoRR}, abs/2307.16092, 2023{\natexlab{b}}.
\newblock \doi{10.48550/ARXIV.2307.16092}.

\bibitem[Rossi et~al.(2023)Rossi, Charpentier, Giovanni, Frasca, G{\"{u}}nnemann, and Bronstein]{rossi2023edge}
Emanuele Rossi, Bertrand Charpentier, Francesco~Di Giovanni, Fabrizio Frasca, Stephan G{\"{u}}nnemann, and Michael~M. Bronstein.
\newblock Edge directionality improves learning on heterophilic graphs.
\newblock In Soledad Villar and Benjamin Chamberlain, editors, \emph{Learning on Graphs Conference, 27-30 November 2023, Virtual Event}, volume 231 of \emph{Proceedings of Machine Learning Research}, page~25. {PMLR}, 2023.

\bibitem[Bevilacqua et~al.(2022)Bevilacqua, Frasca, Lim, Srinivasan, Cai, Balamurugan, Bronstein, and Maron]{bevilacqua2022equivariant}
Beatrice Bevilacqua, Fabrizio Frasca, Derek Lim, Balasubramaniam Srinivasan, Chen Cai, Gopinath Balamurugan, Michael~M. Bronstein, and Haggai Maron.
\newblock Equivariant subgraph aggregation networks.
\newblock In \emph{The Tenth International Conference on Learning Representations, {ICLR} 2022, Virtual Event, April 25-29, 2022}. OpenReview.net, 2022.

\bibitem[Bevilacqua et~al.(2024)Bevilacqua, Eliasof, Meirom, Ribeiro, and Maron]{bevilacqua2024efficient}
Beatrice Bevilacqua, Moshe Eliasof, Eli~A. Meirom, Bruno Ribeiro, and Haggai Maron.
\newblock Efficient subgraph gnns by learning effective selection policies.
\newblock In \emph{The Twelfth International Conference on Learning Representations, {ICLR} 2024, Vienna, Austria, May 7-11, 2024}. OpenReview.net, 2024.

\bibitem[Kipf and Welling(2017)]{kipf2016semi}
Thomas~N. Kipf and Max Welling.
\newblock Semi-supervised classification with graph convolutional networks.
\newblock In \emph{5th International Conference on Learning Representations, {ICLR} 2017, Toulon, France, April 24-26, 2017, Conference Track Proceedings}. OpenReview.net, 2017.

\bibitem[Sato et~al.(2021)Sato, Yamada, and Kashima]{sato2021random}
Ryoma Sato, Makoto Yamada, and Hisashi Kashima.
\newblock Random features strengthen graph neural networks.
\newblock In Carlotta Demeniconi and Ian Davidson, editors, \emph{Proceedings of the 2021 {SIAM} International Conference on Data Mining, {SDM} 2021, Virtual Event, April 29 - May 1, 2021}, pages 333--341. {SIAM}, 2021.
\newblock \doi{10.1137/1.9781611976700.38}.

\bibitem[Velickovic et~al.(2018)Velickovic, Cucurull, Casanova, Romero, Li{\`{o}}, and Bengio]{velivckovic2018graph}
Petar Velickovic, Guillem Cucurull, Arantxa Casanova, Adriana Romero, Pietro Li{\`{o}}, and Yoshua Bengio.
\newblock Graph attention networks.
\newblock In \emph{6th International Conference on Learning Representations, {ICLR} 2018, Vancouver, BC, Canada, April 30 - May 3, 2018, Conference Track Proceedings}. OpenReview.net, 2018.

\bibitem[Shi et~al.(2021)Shi, Huang, Feng, Zhong, Wang, and Sun]{shi2021masked}
Yunsheng Shi, Zhengjie Huang, Shikun Feng, Hui Zhong, Wenjing Wang, and Yu~Sun.
\newblock Masked label prediction: Unified message passing model for semi-supervised classification.
\newblock In Zhi{-}Hua Zhou, editor, \emph{Proceedings of the Thirtieth International Joint Conference on Artificial Intelligence, {IJCAI} 2021, Virtual Event / Montreal, Canada, 19-27 August 2021}, pages 1548--1554. ijcai.org, 2021.
\newblock \doi{10.24963/IJCAI.2021/214}.

\bibitem[Cormen et~al.(2009)Cormen, Leiserson, Rivest, and Stein]{cormen2022introduction}
Thomas~H. Cormen, Charles~E. Leiserson, Ronald~L. Rivest, and Clifford Stein.
\newblock \emph{Introduction to Algorithms, 3rd Edition}.
\newblock {MIT} Press, 2009.

\bibitem[Garey and Johnson(1979)]{garey1979computers}
M.~R. Garey and David~S. Johnson.
\newblock \emph{Computers and Intractability: {A} Guide to the Theory of NP-Completeness}.
\newblock W. H. Freeman, 1979.

\bibitem[Wells(2012)]{wells2012structural}
Alexander~Frank Wells.
\newblock \emph{Structural inorganic chemistry}.
\newblock Oxford University Press, USA, 2012.

\bibitem[Sterling and Irwin(2015)]{sterling2015zinc}
Teague Sterling and John~J. Irwin.
\newblock {ZINC} 15 - ligand discovery for everyone.
\newblock \emph{J. Chem. Inf. Model.}, 55\penalty0 (11):\penalty0 2324--2337, 2015.
\newblock \doi{10.1021/ACS.JCIM.5B00559}.

\bibitem[Dwivedi et~al.(2023)Dwivedi, Joshi, Luu, Laurent, Bengio, and Bresson]{dwivedi2023benchmarking}
Vijay~Prakash Dwivedi, Chaitanya~K. Joshi, Anh~Tuan Luu, Thomas Laurent, Yoshua Bengio, and Xavier Bresson.
\newblock Benchmarking graph neural networks.
\newblock \emph{J. Mach. Learn. Res.}, 24:\penalty0 43:1--43:48, 2023.

\bibitem[Dwivedi et~al.(2022{\natexlab{a}})Dwivedi, Ramp{\'{a}}sek, Galkin, Parviz, Wolf, Luu, and Beaini]{dwivedi2022LRGB}
Vijay~Prakash Dwivedi, Ladislav Ramp{\'{a}}sek, Michael Galkin, Ali Parviz, Guy Wolf, Anh~Tuan Luu, and Dominique Beaini.
\newblock Long range graph benchmark.
\newblock In Sanmi Koyejo, S.~Mohamed, A.~Agarwal, Danielle Belgrave, K.~Cho, and A.~Oh, editors, \emph{Advances in Neural Information Processing Systems 35: Annual Conference on Neural Information Processing Systems 2022, NeurIPS 2022, New Orleans, LA, USA, November 28 - December 9, 2022}, 2022{\natexlab{a}}.

\bibitem[Loshchilov and Hutter(2019)]{loshchilov2018decoupled}
Ilya Loshchilov and Frank Hutter.
\newblock Decoupled weight decay regularization.
\newblock In \emph{7th International Conference on Learning Representations, {ICLR} 2019, New Orleans, LA, USA, May 6-9, 2019}. OpenReview.net, 2019.

\bibitem[T{\"{o}}nshoff et~al.(2024)T{\"{o}}nshoff, Ritzert, Rosenbluth, and Grohe]{tonshoff2023did}
Jan T{\"{o}}nshoff, Martin Ritzert, Eran Rosenbluth, and Martin Grohe.
\newblock Where did the gap go? reassessing the long-range graph benchmark.
\newblock \emph{Trans. Mach. Learn. Res.}, 2024, 2024.

\bibitem[Xu et~al.(2019)Xu, Hu, Leskovec, and Jegelka]{xu2018how}
Keyulu Xu, Weihua Hu, Jure Leskovec, and Stefanie Jegelka.
\newblock How powerful are graph neural networks?
\newblock In \emph{International Conference on Learning Representations}, 2019.

\bibitem[Xu et~al.(2018)Xu, Li, Tian, Sonobe, Kawarabayashi, and Jegelka]{xu2018representation}
Keyulu Xu, Chengtao Li, Yonglong Tian, Tomohiro Sonobe, Ken{-}ichi Kawarabayashi, and Stefanie Jegelka.
\newblock Representation learning on graphs with jumping knowledge networks.
\newblock In Jennifer~G. Dy and Andreas Krause, editors, \emph{Proceedings of the 35th International Conference on Machine Learning, {ICML} 2018, Stockholmsm{\"{a}}ssan, Stockholm, Sweden, July 10-15, 2018}, volume~80 of \emph{Proceedings of Machine Learning Research}, pages 5449--5458. {PMLR}, 2018.

\bibitem[Kreuzer et~al.(2021)Kreuzer, Beaini, Hamilton, L{\'{e}}tourneau, and Tossou]{kreuzer2021rethinking}
Devin Kreuzer, Dominique Beaini, William~L. Hamilton, Vincent L{\'{e}}tourneau, and Prudencio Tossou.
\newblock Rethinking graph transformers with spectral attention.
\newblock In Marc'Aurelio Ranzato, Alina Beygelzimer, Yann~N. Dauphin, Percy Liang, and Jennifer~Wortman Vaughan, editors, \emph{Advances in Neural Information Processing Systems 34: Annual Conference on Neural Information Processing Systems 2021, NeurIPS 2021, December 6-14, 2021, virtual}, pages 21618--21629, 2021.

\bibitem[Pei et~al.(2020)Pei, Wei, Chang, Lei, and Yang]{Pei2020Geom-GCN}
Hongbin Pei, Bingzhe Wei, Kevin~Chen{-}Chuan Chang, Yu~Lei, and Bo~Yang.
\newblock Geom-gcn: Geometric graph convolutional networks.
\newblock In \emph{8th International Conference on Learning Representations, {ICLR} 2020, Addis Ababa, Ethiopia, April 26-30, 2020}. OpenReview.net, 2020.

\bibitem[Lim et~al.(2021)Lim, Hohne, Li, Huang, Gupta, Bhalerao, and Lim]{lim2021large}
Derek Lim, Felix Hohne, Xiuyu Li, Sijia~Linda Huang, Vaishnavi Gupta, Omkar Bhalerao, and Ser{-}Nam Lim.
\newblock Large scale learning on non-homophilous graphs: New benchmarks and strong simple methods.
\newblock In Marc'Aurelio Ranzato, Alina Beygelzimer, Yann~N. Dauphin, Percy Liang, and Jennifer~Wortman Vaughan, editors, \emph{Advances in Neural Information Processing Systems 34: Annual Conference on Neural Information Processing Systems 2021, NeurIPS 2021, December 6-14, 2021, virtual}, pages 20887--20902, 2021.

\bibitem[Platonov et~al.(2023)Platonov, Kuznedelev, Diskin, Babenko, and Prokhorenkova]{platonov2023a}
Oleg Platonov, Denis Kuznedelev, Michael Diskin, Artem Babenko, and Liudmila Prokhorenkova.
\newblock A critical look at the evaluation of gnns under heterophily: Are we really making progress?
\newblock In \emph{The Eleventh International Conference on Learning Representations, {ICLR} 2023, Kigali, Rwanda, May 1-5, 2023}. OpenReview.net, 2023.

\bibitem[Hamilton et~al.(2017)Hamilton, Ying, and Leskovec]{hamilton2017inductive}
William~L. Hamilton, Zhitao Ying, and Jure Leskovec.
\newblock Inductive representation learning on large graphs.
\newblock In Isabelle Guyon, Ulrike von Luxburg, Samy Bengio, Hanna~M. Wallach, Rob Fergus, S.~V.~N. Vishwanathan, and Roman Garnett, editors, \emph{Advances in Neural Information Processing Systems 30: Annual Conference on Neural Information Processing Systems 2017, December 4-9, 2017, Long Beach, CA, {USA}}, pages 1024--1034, 2017.

\bibitem[Dwivedi et~al.(2022{\natexlab{b}})Dwivedi, Luu, Laurent, Bengio, and Bresson]{dwivedi2022graph}
Vijay~Prakash Dwivedi, Anh~Tuan Luu, Thomas Laurent, Yoshua Bengio, and Xavier Bresson.
\newblock Graph neural networks with learnable structural and positional representations.
\newblock In \emph{The Tenth International Conference on Learning Representations, {ICLR} 2022, Virtual Event, April 25-29, 2022}. OpenReview.net, 2022{\natexlab{b}}.

\bibitem[Fey and Lenssen(2019)]{Fey/Lenssen/2019}
Matthias Fey and Jan~Eric Lenssen.
\newblock Fast graph representation learning with pytorch geometric.
\newblock \emph{CoRR}, abs/1903.02428, 2019.
\newblock URL \url{http://arxiv.org/abs/1903.02428}.

\bibitem[Singh et~al.(2016)Singh, Chaudhary, Dhanda, Bhalla, Usmani, Gautam, Tuknait, Agrawal, Mathur, and Raghava]{singh2016satpdb}
Sandeep Singh, Kumardeep Chaudhary, Sandeep~Kumar Dhanda, Sherry Bhalla, Salman~Sadullah Usmani, Ankur Gautam, Abhishek Tuknait, Piyush Agrawal, Deepika Mathur, and Gajendra P.~S. Raghava.
\newblock Satpdb: a database of structurally annotated therapeutic peptides.
\newblock \emph{Nucleic Acids Res.}, 44\penalty0 (Database-Issue):\penalty0 1119--1126, 2016.
\newblock \doi{10.1093/NAR/GKV1114}.

\bibitem[Rozemberczki et~al.(2021)Rozemberczki, Allen, and Sarkar]{rozemberczki2021multi}
Benedek Rozemberczki, Carl Allen, and Rik Sarkar.
\newblock Multi-scale attributed node embedding.
\newblock \emph{J. Complex Networks}, 9\penalty0 (2), 2021.
\newblock \doi{10.1093/COMNET/CNAB014}.

\bibitem[Hall et~al.(2001)Hall, Jaffe, and Trajtenberg]{hall2001nber}
Bronwyn~H Hall, Adam~B Jaffe, and Manuel Trajtenberg.
\newblock The nber patent citation data file: Lessons, insights and methodological tools, 2001.

\bibitem[Tong et~al.(2020)Tong, Liang, Sun, Li, Rosenblum, and Lim]{tong2020digraph}
Zekun Tong, Yuxuan Liang, Changsheng Sun, Xinke Li, David~S. Rosenblum, and Andrew Lim.
\newblock Digraph inception convolutional networks.
\newblock In Hugo Larochelle, Marc'Aurelio Ranzato, Raia Hadsell, Maria{-}Florina Balcan, and Hsuan{-}Tien Lin, editors, \emph{Advances in Neural Information Processing Systems 33: Annual Conference on Neural Information Processing Systems 2020, NeurIPS 2020, December 6-12, 2020, virtual}, 2020.

\bibitem[Zhang et~al.(2021)Zhang, He, Brugnone, Perlmutter, and Hirn]{zhang2021magnet}
Xitong Zhang, Yixuan He, Nathan Brugnone, Michael Perlmutter, and Matthew~J. Hirn.
\newblock Magnet: {A} neural network for directed graphs.
\newblock In Marc'Aurelio Ranzato, Alina Beygelzimer, Yann~N. Dauphin, Percy Liang, and Jennifer~Wortman Vaughan, editors, \emph{Advances in Neural Information Processing Systems 34: Annual Conference on Neural Information Processing Systems 2021, NeurIPS 2021, December 6-14, 2021, virtual}, pages 27003--27015, 2021.

\bibitem[Zhu et~al.(2020)Zhu, Yan, Zhao, Heimann, Akoglu, and Koutra]{zhu2020beyond}
Jiong Zhu, Yujun Yan, Lingxiao Zhao, Mark Heimann, Leman Akoglu, and Danai Koutra.
\newblock Beyond homophily in graph neural networks: Current limitations and effective designs.
\newblock In Hugo Larochelle, Marc'Aurelio Ranzato, Raia Hadsell, Maria{-}Florina Balcan, and Hsuan{-}Tien Lin, editors, \emph{Advances in Neural Information Processing Systems 33: Annual Conference on Neural Information Processing Systems 2020, NeurIPS 2020, December 6-12, 2020, virtual}, 2020.

\bibitem[Chien et~al.(2021)Chien, Peng, Li, and Milenkovic]{chien2021adaptive}
Eli Chien, Jianhao Peng, Pan Li, and Olgica Milenkovic.
\newblock Adaptive universal generalized pagerank graph neural network.
\newblock In \emph{9th International Conference on Learning Representations, {ICLR} 2021, Virtual Event, Austria, May 3-7, 2021}. OpenReview.net, 2021.

\bibitem[Maurya et~al.(2021)Maurya, Liu, and Murata]{maurya2021improving}
Sunil~Kumar Maurya, Xin Liu, and Tsuyoshi Murata.
\newblock Improving graph neural networks with simple architecture design.
\newblock \emph{CoRR}, abs/2105.07634, 2021.
\newblock URL \url{https://arxiv.org/abs/2105.07634}.

\bibitem[Li et~al.(2022)Li, Zhu, Cheng, Shan, Luo, Li, and Qian]{li2022finding}
Xiang Li, Renyu Zhu, Yao Cheng, Caihua Shan, Siqiang Luo, Dongsheng Li, and Weining Qian.
\newblock Finding global homophily in graph neural networks when meeting heterophily.
\newblock In Kamalika Chaudhuri, Stefanie Jegelka, Le~Song, Csaba Szepesv{\'{a}}ri, Gang Niu, and Sivan Sabato, editors, \emph{International Conference on Machine Learning, {ICML} 2022, 17-23 July 2022, Baltimore, Maryland, {USA}}, volume 162 of \emph{Proceedings of Machine Learning Research}, pages 13242--13256. {PMLR}, 2022.

\bibitem[Yang et~al.(2016)Yang, Cohen, and Salakhutdinov]{yang2016revisiting}
Zhilin Yang, William~W. Cohen, and Ruslan Salakhutdinov.
\newblock Revisiting semi-supervised learning with graph embeddings.
\newblock In Maria{-}Florina Balcan and Kilian~Q. Weinberger, editors, \emph{Proceedings of the 33nd International Conference on Machine Learning, {ICML} 2016, New York City, NY, USA, June 19-24, 2016}, volume~48 of \emph{{JMLR} Workshop and Conference Proceedings}, pages 40--48. JMLR.org, 2016.

\bibitem[Keriven(2022)]{keriven2022not}
Nicolas Keriven.
\newblock Not too little, not too much: a theoretical analysis of graph (over)smoothing.
\newblock In Sanmi Koyejo, S.~Mohamed, A.~Agarwal, Danielle Belgrave, K.~Cho, and A.~Oh, editors, \emph{Advances in Neural Information Processing Systems 35: Annual Conference on Neural Information Processing Systems 2022, NeurIPS 2022, New Orleans, LA, USA, November 28 - December 9, 2022}, 2022.

\end{thebibliography}
% For bibLaTeX users:
% \printbibliography

\appendix
\newpage
\section{Mathematical Details}
\label{sec:proofs}
\subsection{Proof of Theorem 4.4}
\begin{proof}
We rewrite the rank-one matrix $\mathbf{X}=\mathbf{u}\mathbf{v}^T$ for some $\mathbf{u}\in\mathbb{R}^n$ and $\mathbb{R}^d$. 
Each $\mathbf{A}_p\mathbf{X}\mathbf{W}_p=\mathbf{u}_p\mathbf{v}_p^T$ is a rank-one matrix with $\mathbf{u}_p = \mathbf{A}_p\mathbf{u}$ and $\mathbf{v}_p^T = \mathbf{v}^T\mathbf{W}_p$. All $\mathbf{v}_p$ are linearly independent to the set of other $\{\mathbf{v}_1,\dots,\mathbf{v}_l\}\setminus \mathbf{v}_p$, as this holds for almost every $\mathbf{W}_1,\dots,\mathbf{W}_l$ with respect to the Lebesgue measure.

We rewrite $\mathbf{x}_i^\prime$ and $\mathbf{x}_j^\prime$ as 
$\mathbf{x}_i^\prime = \sum_{p=1}^l\mathbf{u}_p[i]\mathbf{v}_p^T$. 
We further express this as a vector-matrix product $\mathbf{x}_i^\prime = \mathbf{u}[i]^T\mathbf{V}$ where $\mathbf{u}[i] = \begin{bmatrix} \mathbf{u}_1[i] & \dots & \mathbf{u}_l[i]\end{bmatrix}$ and $\mathbf{V}$ contains $\mathbf{v}_1,\dots,\mathbf{v}_l$ stacked as rows. 
We observe that $\mathbf{x}_i^\prime$ and $\mathbf{x}_j^\prime$ are linearly independent if and only if $\mathbf{u}[i]$ and $\mathbf{u}[j]$ are linearly independent.

For $\mathbf{u}[i]$ and $\mathbf{u}[j]$ to be linearly dependent, there must exist a scalar $c\in\mathbb{R}$ such that $\mathbf{u}[i] - c\cdot\mathbf{u}[j] = 0$. Thus, we also require element-wise for each $i=1,\dots,l$ to hold that $\mathbf{A}_k[i] - m\cdot\mathbf{A}_k[j]\mathbf{u} = 0$. This is further true when $\mathbf{A}_k[i] - m\mathbf{A}_k[j] = 0$ for a.e. $\mathbf{u}$. By our assumption that nodes $v_i$ and $v_j$ are structurally independent, this cannot be satisfied for all elements simultaneously for any $m$.
\end{proof}

\subsection{Proof of Theorem 4.5}
\begin{proof}
Let $\mathbf{d}^{i_1},\dots,\mathbf{d}^{i_l}$ with $i_k\in[n]$ be $l$ linearly independent rows of $\mathbf{D}$. Each $\mathbf{x}_{i_{p}}^\prime$ is linearly independent to the combination of all other $\mathbf{x}_{i_1}^\prime,\dots,\mathbf{x}_{i_{p-1}}^\prime,\dots,\mathbf{x}_{i_{p+1}}^\prime,\dots,\mathbf{x}_{i_l}^\prime$. 

This Theorem then follows Theorem~\ref{theorem:node} by considering a given $\mathbf{x}_{i_p}^\prime$ and replacing $\mathbf{u}_p[j]$ by any combination of the vectors from the other nodes.
By injectivity of $\sigma$, representations remain linearly independent after applying $\sigma$ for a.e. $\mathbf{W}_1,\dots,\mathbf{W}_l$.
\end{proof}

\subsection{Proof of Corollary 5.1}
\begin{proof}
    For all graphs $k$ and all nodes $i$, we have $d_k^i = 1$. Thus, all node pairs are structurally dependent.
\end{proof}

\subsection{Proof of Proposition 5.3}
\begin{proof}
Let $i$ be a root node of $\mathcal{E}_1$ that is not a root node of $\mathcal{E}_2$. Thus, $s_1^i = 0$, while $s_2^i \neq 0$. For any non-root node $j$ of $\mathbf{A}_1$, we have $s_1^j\neq 0$, ensuring structural independence of $i$ and $j$ for any $s_2^j$. 
\end{proof}

\section{Experimental Details}
\label{ap:exp}
In this section, we provide additional details regarding our experiments. All experiments were run on an internal cluster and separately on H100 GPUs, each on a single H100 GPU and an Intel Xeon 8468 Sapphire CPU.

\subsection{MRS-MPNNs}
\paragraph{MRS-SAGE}
\label{sec:model_details}
We define the MRS-SAGE version of the SAGE convolution~\citep{hamilton2017inductive} as
\begin{equation}
    \left[\textrm{MRS-SAGE}\left(\mathbf{X}, \mathcal{E}, f\right)\right]_i = \mathbf{W}\mathbf{x}_i + \sum_{j\in N_i} \frac{1}{d_i} \mathbf{W}_{f(v_i,v_j)}\mathbf{x}_j\, ,
\end{equation}
where $\mathbf{W}\in\mathbb{R}^{d\times d^\prime}$ is the additional feature transformation of the previous state $\mathbf{x}_i$.
\paragraph{MRS-GAT}
For GAT~\citep{velivckovic2018graph}, the corresponding MRS form is
\begin{equation}
    \left[\textrm{MRS-GAT}(\mathbf{X},\mathcal{E})\right]_i = \sum_{j\in N_i} \alpha_{ij} \mathbf{W}_{f(v_i,v_j)}\mathbf{x}_j
\end{equation}
where $\alpha_{ij} = \frac{\textrm{exp}(\textrm{LeakyReLU}(\mathbf{a}^T [\mathbf{W}_{f(v_i,v_j)}||\mathbf{W}_{f(v_i,v_j)}]))}{\sum_{k\in N_i}\textrm{exp}(\textrm{LeakyReLU}(\mathbf{a}^T [\mathbf{W}_{f(v_i,v_j)}||\mathbf{W}_{f(v_i,v_j)}]))}$ with $\mathbf{a}\in\mathbb{R}^{d^\prime}$. In all experiments, GAT and MRS-GAT utilize two heads.
\paragraph{MRS-GIN}
As the transformation in GIN is applied after the aggregation and combination with the previous state, we apply a different GIN instantiation on each edge set: 
\begin{equation}
    \textrm{MRS-GIN}_\textrm{DA}(\mathbf{X},\mathcal{E}) = \textrm{GIN}(\mathbf{X},\mathcal{E}_1)+\textrm{GIN}(\mathbf{X},\mathcal{E}_2)+\textrm{GIN}(\mathbf{X},\mathcal{E}_3)\, .
\end{equation}
\paragraph{MRS-GatedGCN}
We adapt GatedGCN~\citep{bresson2017residual} and its implementation given by \citet{dwivedi2022graph}:
\begin{equation}
    \left[\textrm{MRS-GatedGCN}(\mathbf{X},\mathcal{E})\right]_i = \mathbf{A}\mathbf{x}_i + \sum \frac{\sigma(\mathbf{D}\mathbf{x}_i + \mathbf{E}\mathbf{x}_j + \mathbf{C}\mathbf{e}_{ij}) \odot \mathbf{B}_{f(v_i,v_j)}\mathbf{x}_j}{\sum_{j\in N_i}\sigma(\mathbf{D}\mathbf{x}_i + \mathbf{E}\mathbf{x}_j + \mathbf{C}\mathbf{e}_{ij}) + \epsilon}
\end{equation}
where $\mathbf{A,C,D,E}\in\mathbb{R}^{d\times d^\prime}$ and $\mathbf{B_1},\dots,\mathbf{B_l}\in\mathbb{R}^{d\times d}$ are linear transformations, $\mathbf{e}_{ij}\in\mathbb{R}^d$ are edge attributes, $\sigma$ is the sigmoid function, and $\epsilon=1e-6$ is a small constant.

\subsection{Improving the Learning Process}
Our implementation is built on the Long Range Graph Benchmark (LRGB)~\citep{dwivedi2022LRGB,tonshoff2023did} which is available under the MIT license. It is based on PyTorch Geometric~\citep{Fey/Lenssen/2019}. We add our models without making changes to the optimization and data construction parts. 

\subsubsection{Models and Optimization}
All models perform $k$ iterations of message-passing, with ReLU as a non-linear activation function.
We reuse the standard optimization process from \citep{dwivedi2022LRGB} and use the AdamW~\citep{loshchilov2018decoupled} optimizer with a cosine learning rate schedule. The cross-entropy loss is used for optimization, and the average precision (AP) as a metric. All models use at most $\num{500000}$ parameters to ensure fairness.

\subsubsection{Datasets}
\paragraph{ZINC}
We utilize the ZINC dataset~\citep{sterling2015zinc} that contains \num{250000} chemical compounds that are represented as graphs. We also utilize the ZINC12k subset that contains a subset of $\num{12000}$ chemical compounds. Each node represents an atom, and each edge is a bond between two atoms. On average, a graph has around $23$ nodes and $50$ edges. Node features are given as a single value indicating its corresponding type of heavy atom. We do not utilize edge features. The objective is given as a graph regression task, which corresponds to the prediction of the constrained solubility of the molecule. The mean absolute error (MAE) is used as the loss function and for the score. Each experiment on ZINC is repeated for three random seeds. ZINC is freely commercially available under the license DbCL.

\paragraph{Peptides-Func}
The Peptides dataset consists of \num{15535} peptides, which are short molecular chains~\citep{singh2016satpdb}. As with ZINC, nodes of the graph represent atoms and edges the bonds between them. They are part of the LRGB as peptides have a large diameter while each node has a low average degree of around $2$. Thus, it is argued that this dataset requires models that can combine distant information in the graph, i.e., models with many layers. Node and edge features are constructed using molecular SMILES based on the atom types. This dataset was released under license CC BY-NC 4.0.

The task is to predict the molecular properties of each peptide, i.e., a multi-label graph classification task. Each graph belongs, on average, to $1.65$ of $10$ classes. 
As proposed by \citet{dwivedi2022LRGB}, the cross-entropy (CE) loss is used for optimization, and the unweighted mean average precision (AP) as the metric. We follow their same data split, i.e., $70\%$ for training and $15\%$ for validation and testing.

 The resulting representations are globally aggregated using the mean and mapped to class probabilities using a linear layer. 

\paragraph{Peptides-Struct}
The dataset is the same as used for Peptides-Func. The task is to predict five continuous geometric properties of the peptides, i.e., a multi-label graph regression task. The same data split is utilized. The MAE is used for both optimization and as the target metric.

\subsubsection{Orderings}
For all orderings, we compute a single value $r_i$ per node $v_i$ and construct the strict partial ordering $i\prec j \Leftrightarrow r_i < r_j$. 
\paragraph{Random} For the random ordering, we assign each node $v_i\in\mathcal{V}$ a unique random index $r_i\in [0,|\mathcal{V}|]$. This index is consistent across layers and epochs. The edges within each computational graph will be similar to those from different computational graphs. Thus, we expect the optimal solution to be achieved when all transformations are equal.  
\paragraph{Features} This ordering utilizes the initial node features $\mathbf{x}_i\in\mathbb{R}^d$ by summing $r_i = \sum_{c\in[d]} \mathbf{x}_{ic}$ over all features of that node.
\paragraph{PPR} Here, we perform $15$ iterations of Personalized PageRank (PPR) with a restart probability $\alpha = 0.1$. This provides a finer node centrality measure as opposed to the node degree. A finer ordering will have fewer edges in the third computational graph, but the similarity between edges in each computational graph may be lowered. A node's role in a graph is connected to its centrality, e.g., influential persons in social networks have many connections.
\paragraph{Degree} As a coarser node centrality measure, we consider the node degree $r_i = d_i$. For molecular graphs, the node degree is closely connected with its role within the molecule. The degree is also much more efficient to compute compared to PPR.

\subsubsection{Comparing Partial Orderings (Table 1)}
\begin{table*}[t]
\caption{Best hyperparameters for results in Table~\ref{tab:ordering}}
\label{tab:ordering_hyper}
\vskip 0.15in
\begin{center}
\begin{small}
\begin{sc}
\begin{tabular}{lccccccc}
\toprule
Dataset & & \multicolumn{2}{c}{ZINC} \\
Parameter & Split & LR & Layers\\
\midrule
\multirow{2}{*}{GCN} & Train & $0.001$ & $8$ \\
& Val & $0.0003$ & $8$ \\
\multirow{2}{*}{$\textrm{MRS-GCN}_\textrm{DA}$ (random)} & Train & $0.001$ & $8$ \\
& Val & $0.003$ & $16$ \\
\multirow{2}{*}{$\textrm{MRS-GCN}_\textrm{DA}$ (Features)} & Train & $0.001$ & $8$ \\
& Val & $0.001$ & $8$ \\
\multirow{2}{*}{$\textrm{MRS-GCN}_\textrm{DA}$ (PPR)} & Train & $0.001$ & $8$ \\
& Val & $0.001$ & $8$ \\
\multirow{2}{*}{$\textrm{MRS-GCN}_\textrm{DA}$ (Degree)} & Train & $0.0003$ & $8$ \\
& Val & $0.0003$ & $8$ \\
\bottomrule
\end{tabular}
\end{sc}
\end{small}
\end{center}
\vskip -0.1in
\end{table*}
\paragraph{Experimental Setup}
To compare the performances of the partial orderings we considered, we evaluate them on the ZINC task. All models have a fixed hidden dimension of $64$. We tune the learning rate for all models with values $\in\{0.003,0.001,0.0003\}$ and the number of layers $\in\{1,2,4,8,16,32\}$. All experiments are repeated for three random seeds. Best hyperparameters are presented in Table~\ref{tab:ordering_hyper}. 
We compute the ordering and the split once in each forward pass.
The symmetric normalization of the adjacency matrix is performed once per forward pass for GCN and MRS-GCN. Displayed runtimes are for eight-layer models. These experiments run for a total of around $60$ hours on one H100 GPU. 

\paragraph{Runtime}
We provide an additional in-depth runtime analysis in Table~\ref{tab:runtime}. We measure the time that applying the steps for performing normalization, ordering, splitting, applying feature transformations, and aggregating messages take. Normalization, ordering, and splitting are fixed throughout training, so they only have to be calculated once. As the same total number of parameters is used for all models, applying the three transformations $\mathbf{W}_1,\mathbf{W}_2,\mathbf{W}_3$ only takes around $25\%$ more time. Aggregating messages from all computational graphs takes an additional time of around $15\%$. In total, the differences are slightly lower, as all other computations are equivalent.

\begin{table*}[t]
\caption{Execution times in milliseconds for applying each step a single time on a batch of 32 graphs of the ZINC dataset. Normalization refers to applying symmetric degree normalization on the adjacency matrix, ordering refers to calculating a single scalar per node, and splitting refers to assigning each edge to one of the three edge sets based on the ordering. Transformation describes applying $\mathbf{W}$ or $\mathbf{W}_1,\mathbf{W}_2,\mathbf{W}_3$ on the node state, and aggregation means collecting the weighted sum over all utilized edge relations.}
\label{tab:runtime}
\vskip 0.15in
\begin{center}
\scriptsize
\begin{sc}
\begin{tabular}{lccccccc}
\toprule
Method & Normalization & Ordering & Splitting & Transformation & Aggregation \\
\midrule
GCN & $0.129$ & - & - & $0.044$ & $0.091$ \\
MRS-GCN$_\textrm{DA}$ {\tiny (random)} & 0.130 & 0.019 & 0.249 & 0.051 & 0.097 \\
MRS-GCN$_\textrm{DA}$ {\tiny (Features)} & 0.131 & 0.022 & 0.268 & 0.054 & 0.106 \\
MRS-GCN$_\textrm{DA}$ {\tiny (PPR)} & 0.128 & 1.150 & 0.212 & 0.055 & 0.092 \\
MRS-GCN$_\textrm{DA}$ {\tiny (degree)} & 0.128 & 0.042 & 0.261 & 0.055 & 0.104 \\
\bottomrule
\end{tabular}
\end{sc}
\end{center}
\vskip -0.1in
\end{table*}

\subsubsection{Preventing Rank Collapse (Figure 2)}
We utilize random graphs of the ZINC dataset. We perform a single linear transformation to change the feature dimension to $16$. We then apply message-passing iterations, each followed by a ReLU activation. Each iteration maintains the feature dimension as $16$. Feature transformations are not shared between iterations. The Rank-one distance is defined as
\begin{equation}
\mathrm{ROD}(\mathbf{X}) = \left\| \frac{\mathbf{X}}{\|\mathbf{X}\|} - \frac{\mathbf{uv}^T}{\|\mathbf{uv}^T\|}\right\|
\end{equation}
where $\mathbf{u}\in\mathbb{R}^n$ is the column $\mathbf{v}\in\mathbb{R}^d$ the row of $\mathbf{X}$ with the largest norm~\citep{roth2024simplifying}. All norms are nuclear norms. As this metric generalizes the Dirichlet energy~\citep{cai2020note}, constructing models that keep ROD constant, also prevent over-smoothing.
It is calculated after the ReLU activation. We repeat this experiment for $50$ random seeds. These experiments run for less than one minute on a CPU.

\subsubsection{Details on Table 2}

\begin{table*}[t]
\caption{Best hyperparameters for results in Table~\ref{tab:zinc_full}}
\label{tab:overview_hyper}
\vskip 0.15in
\begin{center}
\begin{small}
\begin{sc}
\begin{tabular}{lccccccc}
\toprule
Dataset & \multicolumn{2}{c}{ZINC (Train)} & \multicolumn{2}{c}{ZINC (Test)} \\
Parameter & LR & Layers & LR & Layers\\
\midrule
GCN & $0.0003$ & $16$ & $0.0003$ & $16$\\
$\textrm{MRS-GCN}_{\textrm{DA}}$ & $0.0003$ & $8$ & $0.0001$ & $8$\\
\midrule
SAGE & $0.0003$ & $8$ & $0.0003$ & $8$\\
$\textrm{MRS-SAGE}_{\textrm{DA}}$ & $0.0003$ & $8$ & $0.0003$ & $8$\\
\midrule
GAT & $0.0003$ & $8$ & $0.0003$ & $8$\\
GAT+degree & $0.0003$ & $8$ & $0.0003$ & $8$\\
$\textrm{MRS-GAT}_{\textrm{DA}}$ & $0.0003$ & $8$ & $0.0003$ & $8$\\
\midrule
GIN & $0.0003$ & $8$ & $0.0003$ & $8$\\
$\textrm{MRS-GIN}_{\textrm{DA}}$ & $0.0003$ & $8$ & $0.0001$ & $8$\\
\midrule
GatedGCN & $0.0003$ & $8$ & $0.0003$ & $8$\\
$\textrm{MRS-GatedGCN}_{\textrm{DA}}$ & $0.0003$ & $8$ & $0.0003$ & $8$\\
\bottomrule
\end{tabular}
\end{sc}
\end{small}
\end{center}
\vskip -0.1in
\end{table*}

\begin{table*}[t]
\caption{Hidden dimensions and total parameter count for eight-layer models on ZINC and ZINC12k.}
\label{tab:parameter_count}
\vskip 0.15in
\begin{center}
\begin{small}
\begin{sc}
\begin{tabular}{lcc}
\toprule
Dataset & Hidden Dimension & Parameters \\
\midrule
GCN & $247$ & $498200$\\
$\textrm{MRS-GCN}_{\textrm{DA}}$ & $143$ & $496656$\\
\midrule
SAGE & $175$ & $497176$\\
$\textrm{MRS-SAGE}_{\textrm{DA}}$ & $124$ & $497117$\\
\midrule
GAT & $174$ & $497119$\\
GAT+degree & $174$ & $497151$\\
$\textrm{MRS-GAT}_{\textrm{DA}}$ & $101$ & $497022$\\
\midrule
GIN & $143$ & $498928$\\
$\textrm{MRS-GIN}_{\textrm{DA}}$ & $101$ & $497830$\\
\midrule
GatedGCN & $110$ & $495551$\\
$\textrm{MRS-GatedGCN}_{\textrm{DA}}$ & $82$ & $495363$\\
\bottomrule
\end{tabular}
\end{sc}
\end{small}
\end{center}
\vskip -0.1in
\end{table*}

For ZINC, we tune the number of layers in $\{1,2,4,8,16,32\}$ and the base learning rate in $\{0.001,0.0003,0.0001\}$. All runs utilize a cosine learning rate schedule with a maximum of $500$ epochs for ZINC. As given by our reference implementation~\citep{dwivedi2022graph}, a batch size of $32$ is used. The hidden dimension of each model is reduced until less than $\num{500000}$ parameters are used. As an example, the hidden dimensions of the eight-layer models are shown in Table~\ref{tab:parameter_count}. Best hyperparameters for each experiment are displayed in Table~\ref{tab:overview_hyper}. 
These experiments require around $4500$ hours on an H100 GPU.

For Peptides-Func, we tune the number of layers in $\{1,2,4,8,16,32\}$ and the base learning rate in $\{0.005,0.001,0.0005\}$ using a grid search. As given by our reference implementation~\citep{tonshoff2023did}, a batch size of $200$ is used.
Best hyperparameters for each experiment are displayed in Table~\ref{tab:func_hyper}. These experiments require around $70$ hours on an H100 GPU. Results are presented in Table~\ref{tab:peptides_func}.
\begin{table*}[tb]
\caption{Mean MAE and standard deviations over three runs on the Peptides-Func dataset (pairwise best results marked in \textbf{bold}). The learning rate and the number of layers are tuned. Best hyperparameters are provided in brackets. Train MAE and AP are the overall minimum, while test AP is based on the best validation AP.}
\label{tab:peptides_func}
\vskip 0.15in
\begin{center}
\begin{small}
\begin{sc}
\begin{tabular}{lccccc}
\toprule
\multirow{2}{*}{Method} & \multicolumn{3}{c}{Peptides-Func} & \multicolumn{2}{c}{Peptides-Struct} \\ & Train (MAE) & Train (AP) & Test (AP) & Train (MAE) & Test (MAE)\\
\midrule
GCN& $0.033\pm0.023$ & $0.989\pm0.004$ & $0.574\pm0.005$ & $0.146\pm0.032$ & $0.303\pm0.009$\\
$\textrm{MRS-GCN}_{\textrm{DA}}$ & $\mathbf{0.003}\pm0.000$ & $\mathbf{0.999}\pm0.000$ & $\mathbf{0.588}\pm0.011$ & $\mathbf{0.052}\pm0.004$ & $\mathbf{0.301}\pm0.007$\\
\midrule
SAGE & $0.011\pm0.004$ & $\mathbf{0.999}\pm0.000$ & $0.566\pm0.011$ & $0.168\pm0.004$ & $0.317\pm0.0001$\\
$\textrm{MRS-SAGE}_{\textrm{DA}}$ & $\mathbf{0.002}\pm0.000$ & $\mathbf{0.999}\pm0.000$ & $\mathbf{0.615}\pm0.008$ & $\mathbf{0.055}\pm0.004$ & $\mathbf{0.293}\pm0.000$\\
\midrule
GIN & $0.031\pm0.021$ & $0.980\pm0.021$ & $0.553\pm0.016$ & $0.228\pm0.012$ & $0.307\pm0.001$\\
$\textrm{MRS-GIN}_{\textrm{DA}}$ & $\mathbf{0.007}\pm0.002$ & $\mathbf{0.999}\pm0.000$ & $\mathbf{0.576}\pm0.007$ & $\mathbf{0.149}\pm0.013$ & $\mathbf{0.301}\pm0.002$ \\
\bottomrule
\end{tabular}
\end{sc}
\end{small}
\end{center}
\vskip -0.1in
\end{table*}

\begin{table*}[tb]
\caption{Best hyperparameters for the results in Table~\ref{tab:peptides_func}. Entries are formatted as optimal learning rate / number of layers.}
\label{tab:func_hyper}
\vskip 0.15in
\begin{center}
\begin{small}
\begin{sc}
\begin{tabular}{lccccc}
\toprule
\multirow{2}{*}{Method} & \multicolumn{3}{c}{Peptides-Func} & \multicolumn{2}{c}{Peptides-Struct} \\ & Train (MAE) & Train (AP) & Test (AP) & Train (MAE) & Test (MAE)\\
\midrule
GCN& 0.005 / 4 & 0.005 / 4 & 0.001 / 8 & 0.005 / 8 & 0.0005 / 32 \\
$\textrm{MRS-GCN}_{\textrm{DA}}$ & 0.005 / 16 & 0.005 / 16 & 0.001 / 16 & 0.005 / 8 & 0.001 / 32\\
\midrule
SAGE & 0.001 / 16 & 0.001 / 16 & 0.001 / 8 & 0.005 / 8 & 0.0005 / 16\\
$\textrm{MRS-SAGE}_{\textrm{DA}}$ & 0.001 / 16 & 0.001 / 16 & 0.001 / 16 & 0.005 / 8 & 0.0005 / 32\\
\midrule
GIN & 0.0001 / 8 & 0.0001 / 8 & 0.001 / 8 & 0.001 / 8 & 0.0005 / 16\\
$\textrm{MRS-GIN}_{\textrm{DA}}$ & 0.001 / 8 & 0.001 / 8 & 0.0005 / 16 & 0.001 / 8 & 0.0005 / 32\\
\bottomrule
\end{tabular}
\end{sc}
\end{small}
\end{center}
\vskip -0.1in
\end{table*}

\subsubsection{Details on Figure 3}
We track the training loss on the full ZINC dataset. Hyperparameters are selected based on the minimal achieved final training loss for each model. We repeated all settings for three random seeds. All runs utilize a cosine learning rate schedule with a maximum of $500$ epochs. As given by~\citet{dwivedi2022graph}, a batch size of $32$ is used. Including hyperparameter optimization, these experiments took $2700$ hours on an H100 GPU.

\subsubsection{Details on Table 3}
%ZINC
\begin{table*}[t]
\caption{Standard deviations for the results in Table~\ref{tab:layer}}
\label{tab:stdev}
\vskip 0.15in
\begin{center}
\footnotesize
\begin{tabular}{lcccccc}
\toprule
Method & 1 & 2 & 4 & 8 & 16 & 32 \\
\midrule
GCN & $0.002$ & $0.001$ & $0.004$ & $0.011$ & $0.014$ & $0.002$ \\
$\textrm{MRS-GCN}_{\textrm{DA}}$ & $0.002$ & $0.007$ & $0.011$ & $0.003$ & $0.014$ & $0.006$ \\
\midrule
GCN + Res & $0.003$ & $0.004$ & $0.002$ & $0.014$ & $0.021$ & $0.019$ \\
$\textrm{MRS-GCN}_{\textrm{DA}}$ + Res & $0.001$ & $0.001$ & $0.011$ & $0.002$ & $0.003$ & $0.004$\\
\midrule
GCN + JK & $0.004$ & $0.002$ & $0.003$ & $0.011$ & $0.009$ & $0.010$\\
$\textrm{MRS-GCN}_{\textrm{DA}}$ + JK & $0.004$ & $0.000$ & $0.003$ & $0.003$ & $0.005$ & $0.009$ \\
\midrule
GCN + LapPE & $0.017$ & $0.008$ & $0.005$ & $0.018$ & $0.031$ & $0.013$ \\
$\textrm{MRS-GCN}_{\textrm{DA}}$ + LapPE & $0.020$ & $0.060$ & $0.014$ & $0.020$ & $0.009$ & $0.002$ \\
\bottomrule
\end{tabular}
\end{center}
\vskip -0.1in
\end{table*}

\begin{table*}[t]
\caption{Best hyperparameters for results in Table~\ref{tab:layer}}
\label{tab:layer_hyper}
\vskip 0.15in
\begin{center}
\begin{small}
\begin{sc}
\begin{tabular}{lcccccc}
\toprule
Method & 1 & 2 & 4 & 8 & 16 & 32 \\
\midrule
GCN & $0.0003$ & $0.0003$ & $0.0003$ & $0.0003$ & $0.0003$ & $0.001$ \\
$\textrm{MRS-GCN}_\textrm{DA}$ & $0.0003$ & $0.0003$ & $0.0003$ & $0.0003$ & $0.0003$ & $0.0003$ \\
\midrule
GCN + Res & $0.0003$ & $0.0003$ & $0.0003$ & $0.001$ & $0.001$ & $0.001$ \\
$\textrm{MRS-GCN}_\textrm{DA}$ + Res & $0.0003$ & $0.0003$ & $0.0003$ & $0.001$ & $0.001$ & $0.003$ \\
\midrule
GCN + JK & $0.0003$ & $0.0003$ & $0.0003$ & $0.0003$ & $0.0003$ & $0.0003$ \\
$\textrm{MRS-GCN}_\textrm{DA}$ + JK & $0.0003$ & $0.0003$ & $0.001$ & $0.0003$ & $0.0003$ & $0.0003$ \\
\midrule
GCN + LapPE & $0.001$ & $0.0003$ & $0.0003$ & $0.0003$ & $0.001$ & $0.0003$ \\
$\textrm{MRS-GCN}_\textrm{DA}$ + LapPE & $0.001$ & $0.003$ & $0.001$ & $0.0003$ & $0.0003$ & $0.0003$ \\
%\midrule
%GCN + RWSE & $0.005$ & $0.005$ & $0.005$ & $0.005$ & $0.001$ & $0.001$ \\
%$\textrm{MRS-GCN}_\textrm{DA}$ + RWSE & $0.0005$ & $0.001$ & $0.001$ & $0.001$ & $0.001$ & $0.0005$\\
\bottomrule
\end{tabular}
\end{sc}
\end{small}
\end{center}
\vskip -0.1in
\end{table*}

\paragraph{Jumping Knowledge}
We store the output of each layer after applying the non-linearity into a set $\mathcal{X} = \{ \mathbf{X}^{(1)},\dots,\mathbf{X}^{(k)}\}$. The resulting representation is then obtained by either taking the element-wise maximum, or concatenating all representation per node. This output is used instead of the output of the last message-passing layer.

\paragraph{Residual Connections}
This technique adds the previous state to the message-passing operation's output for each layer after the non-linearity.

\paragraph{Laplacian Positional Encoding}
We obtain Laplacian positional encodings based on eight frequencies and concatenate these features to the initial node representations.

\paragraph{Optimization}
For each model type and layer count, we tune the learning rate in $\{0.001,0.0003,0.0001\}$. Reported test scores are based on the epoch of the best validation score. All runs utilize a cosine learning rate schedule with a maximum of $400$ epochs. 
As given by~\citet{tonshoff2023did}, a batch size of $32$ is used.
These experiments require a total of around $110$ hours on an H100 GPU.

\subsection{Comparison with State-of-the-Art}
Our models are integrated into the implementation of Dir-GNNs~\citep{rossi2023edge}. This implementation is available under the MIT license.

\subsection{Memory Costs}
\label{sec:memory}
We present the empirical memory costs in Table~\ref{tab:memory}. As the GCN and the $\textrm{MRS-GCN}_\textrm{DA}$ utilize the same number of parameters, their memory consumption is almost identical.
\begin{table*}[t]
\caption{Memory usage during training of ZINC12k. Values are averaged over five runs.}
\label{tab:memory}
\vskip 0.15in
\begin{center}
\begin{small}
\begin{sc}
\begin{tabular}{lcccccc}
\toprule
Method & GPU Memory & Main Memory \\
\midrule
GCN & $1.918$ GB & $1.218$ GB \\
$\textrm{MRS-GCN}_\textrm{DA}$ & $1.914$ GB & $1.265$ GB \\
\bottomrule
\end{tabular}
\end{sc}
\end{small}
\end{center}
\vskip -0.1in
\end{table*}

\subsubsection{Datasets}
\paragraph{Chameleon and Squirrel}
These two datasets are based on pages about particular topics in Wikipedia. Nodes represent articles on that topic and edges their links. Node features are constructed as the appearance of particular nouns~\citep{rozemberczki2021multi}. The task is to classify each article based on their average monthly traffic~\cite{Pei2020Geom-GCN}. Chameleon consists of \num{2277} nodes and \num{36101} edges. Squirrel consists of \num{5201} nodes and \num{217073} edges. To the best of our knowledge, the dataset was released without a license.

\paragraph{Arxiv-Year}
In this dataset, nodes correspond to publications, and an edge is constructed when a publication cites another. The task is to classify the publication year into one of five time spans. It consists of \num{169343} nodes and \num{1166243} edges. Nodes are given by word $128$-dimensional embeddings of the title and abstract of the corresponding publication. Arxiv-Year is released under the ODC-BY license.

\paragraph{Snap-Patents}
Nodes correspond to patents, for which the year it was granted should be classified into one of five time spans. Edges similarly correspond to citations between patents. This dataset consists of \num{2923922} nodes and \num{13975791} edges. This dataset was released without a license by \citet{hall2001nber}.

\subsubsection{Implementational Details}
\begin{table*}[t]
\caption{Best hyperparameters for the results in Table~\ref{tab:hetero}.}
\label{tab:hetero_hyper}
\vskip 0.15in
\begin{center}
\begin{small}
\begin{sc}
\begin{tabular}{lccc}
\toprule
Dataset & Learning Rate & Layers & Dropout Ratio \\
\midrule
Chameleon & $0.005$ & $6$ & $0.2$ \\
Squirrel & $0.001$ & $6$ & $0.0$ \\
Roman-Empire & $0.005$ & $6$ & $0.2$ \\
Arxiv-Year & $0.005$ & $5$ & $0.6$ \\
Snap-Patents & $0.01$ & $6$ & $0.0$ \\
\bottomrule
\end{tabular}
\end{sc}
\end{small}
\end{center}
\vskip -0.1in
\end{table*}

We compare to several state-of-the-art methods for directed graphs, namely DiGCN~\citep{tong2020digraph}, MagNet~\citep{zhang2021magnet}, and Dir-GNN~\citep{rossi2023edge}, and state-of-the-art methods for heterophilic graphs, namely $\textrm{H}_2\textrm{GCN}$~\citep{zhu2020beyond}, GPR-GNN~\citep{chien2021adaptive}, LINKX~\citep{lim2021large}, FSGNN~\citep{maurya2021improving}, ACM-GCN~\citep{luan2022revisiting}, GloGNN~\citep{li2022finding}, and Gradient Gating~\citep{rusch2022gradient}.

We use the same model as \citet{rossi2023edge}, with the only change being the exchange of each MPNN module with the corresponding $\textrm{MRS-MPNN}_\textrm{DA}$ . The models consist of $k$ layers of message-passing, each followed by ReLU and potentially Dropout. All representations are normalized by the $L2$-norm $\|\mathbf{X}\|_2$. Final representations are obtained by Knowledge Knowledge, either concatenating all intermediate states (cat) or taking the element-wise maximum (max). Node degree is used as ordering for all experiments.
We tune $\textrm{MRS-Dir-GNN}_{\textrm{DA}}$ using the same hyperparameters and their ranges as performed for Dir-GNN using the same implementation: The learning rate $\in \{0.01,0.005,0.001,0.0005\}$, number of layers $\in \{4,5,6\}$, jumping knowledge $\in\{\textrm{cat},\textrm{max}\}$, dropout $\in\{0.0,0.2,0.4,0.6\}$.
We use their optimal values for hidden feature dimension $\in\{32,64,128,256,512\}$, normalization $\in\{\textrm{True}, \textrm{False}\}$ and $\alpha\in\{0.,0.5,1\}$ for each task. As given in their implementation, the patience of stopping training based on not improving the validation accuracy is set to $200$ for Roman-Empire, Snap-Patents, and Arxiv-Year and to $400$ for Squirrel and Chameleon. Consequently, $\textrm{MRS-SAGE}_\textrm{DA}$ is used for Roman-Empire and $\textrm{MRS-GCN}_\textrm{DA}$ for the other datasets. The best-performing hyperparameters are presented in Table~\ref{tab:hetero_hyper}.

\subsubsection{Runtime}
On an H100 GPU, each run for Chameleon takes around $10$ seconds, for Squirrel $20$ seconds, for Roman-Empire $90$ seconds, for Arxiv-Year $15$ minutes, and for Snap-Patents $30$ minutes. In total, these Experiments take around $150$ GPU hours.
We estimate the time of our preliminary experiments with various versions of MRS-MPNNs to additional \num{1500} hours.

\subsection{Additional Experiments on Homophilic Datasets}

\begin{table*}[t]
\caption{Test accuracies for homophilic datasets.}
\label{tab:homo}
\vskip 0.15in
\begin{center}
\begin{small}
\begin{sc}
\begin{tabular}{lccc}
\toprule
Method & Cora (Acc) & CiteSeer (Acc) & PubMed (Acc) \\
\midrule
GCN & $81.1\pm0.9$ & $66.6\pm1.0$ & $75.5\pm1.2$ \\
$\textrm{MRS-GCN}_\textrm{DA}$ & $80.0\pm1.2$ & $62.8\pm1.7$ & $76.2\pm0.7$ \\
\bottomrule
\end{tabular}
\end{sc}
\end{small}
\end{center}
\vskip -0.1in
\end{table*}

\begin{table*}[t]
\caption{Best hyperparameters for Table~\ref{tab:homo}. Entries are formatted as Learning rate / dropout ratio / number of layers.}
\label{tab:homo_hyper}
\vskip 0.15in
\begin{center}
\begin{small}
\begin{sc}
\begin{tabular}{lccc}
\toprule
Method & Cora (Acc) & CiteSeer (Acc) & PubMed (Acc) \\
\midrule
GCN & $0.01 / 0.0 / 2$ & $0.03 / 0.6 / 2$ & $0.01 / 0.6 / 4$ \\
$\textrm{MRS-GCN}_\textrm{DA}$ & $0.03 / 0.6 / 2$ & $0.03 / 0.4 / 2$ & $0.01 / 0.4 / 4$ \\
\bottomrule
\end{tabular}
\end{sc}
\end{small}
\end{center}
\vskip -0.1in
\end{table*}

We additionally conducted experiments on three datasets for homophilic node classification, namely Cora, CiteSeer, and PubMed~\citep{yang2016revisiting}. We evaluate the GCN and the $\textrm{MRS-GCN}_\textrm{DA}$. We tune the learning rate in $\{0.03,0.01,0.003\}$, the dropout ratio in $\{0.0,0.2,0.4,0.6,0.8\}$, and the number of layers in $\{1,2,4,8,16,32\}$ for both methods. These experiments take around three hours on a single H100 GPU. 

Results are presented in Table~\ref{tab:homo}. The accuracy achieved by $\textrm{MRS-GCN}_\textrm{DA}$ is lower for Cora and CiteSeer, which indicates that splitting edges based on the degree does not hold relevant information for this task. Instead, these homophilic tasks are known to benefit from smoothing dynamics~\citep{keriven2022not}. Having $\mathbf{W}_1=\mathbf{W}_2=\mathbf{W}_3$ would be better, but this solution is apparently harder to find. Different edge-splitting functions would likely result in a higher accuracy, e.g., by splitting based on inter- and intra-community edges.

\end{document}